\documentclass{article}

  \usepackage[preprint]{neurips_2026}

\usepackage[utf8]{inputenc} 
\usepackage[T1]{fontenc}    
\usepackage{hyperref}       
\usepackage{url}            
\usepackage{booktabs}       
\usepackage{amsfonts}       
\usepackage{nicefrac}       
\usepackage{microtype}      
\usepackage{xcolor}         
\usepackage{amsmath}
\usepackage{graphicx}
\usepackage{tabularx}
\usepackage[capitalize]{cleveref}
\usepackage{makecell}
\usepackage{multirow}
\usepackage{float}
\usepackage[table]{xcolor}
\usepackage{threeparttable}
\usepackage[shortlabels,inline]{enumitem}
\usepackage{wrapfig}
\usepackage{multirow} 
\usepackage{tcolorbox}
\usepackage{subcaption}
\usepackage{caption}
\usepackage[Export]{adjustbox}
\newcommand{\myparagraph}[1]{\noindent\textbf{#1}}

\RequirePackage{xspace}
\makeatletter
\DeclareRobustCommand\onedot{\futurelet\@let@token\@onedot}
\def\@onedot{\ifx\@let@token.\else.\null\fi\xspace}

\def\ie{\emph{i.e}\onedot} 

\def\cf{\emph{cf}\onedot}

\makeatother

\title{Do‑Undo Bench: Reversibility for Action Understanding in Image Generation}

\author{%
  Shweta Mahajan\textsuperscript{1, 2\footnotemark[1]} \quad Shreya Kadambi\textsuperscript{3\footnotemark[1]} \quad  Hoang Le\textsuperscript{3} \quad Rajeev Yasarla\textsuperscript{3} \and  \textbf{Apratim Bhattacharyya\textsuperscript{3} \quad Munawar Hayat\textsuperscript{3} \quad  Fatih Porikli\textsuperscript{3}}\\
 \textsuperscript{1}York University \quad \textsuperscript{2}Vector Institute for AI \quad \textsuperscript{3}Qualcomm AI Research\textsuperscript{\footnotemark[2]}\\
 }
 \definecolor{myPink}{RGB}{255, 105, 180}

\hypersetup{
    colorlinks=true,
    urlcolor=myPink
}
\begin{document}

\maketitle


\begin{abstract}
We introduce the Do-Undo task and benchmark to address a critical gap in vision-language models: understanding and generating plausible scene transformations driven by real-world actions. 
Unlike prior work that relies on prompt-based image generation and editing to perform action-conditioned image manipulation, our training hypothesis requires models to simulate the outcome of a real-world action and then reverse it to the original state. 
This forward–reverse requirement tests genuine cause-and-effect understanding rather than stylistic or semantic edits. 
We curate a high-quality benchmark of reversible actions from real-world scenarios to enable robust action grounding. 
Our experiments reveal that current models struggle with action reversibility, highlighting the need to evaluate action understanding. 
Do-Undo provides an intuitive testbed for evaluating and advancing action-aware generation in multimodal systems that must reason about real-world dynamics.
\end{abstract}
\renewcommand{\thefootnote}{\fnsymbol{footnote}}
\footnotetext[1]{Equal contribution.}
\footnotetext[2]{Qualcomm AI Research is an initiative of Qualcomm Technologies, Inc.}
\footnotetext[3]{Dataset at: \url{https://huggingface.co/datasets/doundo/doundobench}}
\footnotetext[4]{Project page: \url{https://s-mahajan.github.io/Do-Undo-Bench/}}
\section{Introduction}
\label{sec:intro}
Advances in vision-language foundation models (VLMs) have enabled remarkable progress in text-driven image synthesis and editing \cite{deng2025bagel, brooks2023instructpix2pix, OpenAIDocsImageGen, gemini-flash-2025, sheynin2024emu, hui2024hq}, with new capabilities in creative applications and synthetic data generation. Despite these advances, current models remain fundamentally limited in their ability to understand and simulate the physical dynamics of real-world scenes \cite{al2024unibench,meng2024towards,kang2024far,azzolini2025cosmos,li2017visual}. 
Existing approaches focus on object-level manipulations, such as adding or removing objects, while neglecting the underlying cause-and-effect relationships that govern physical interactions \cite{bhattad2025visual,ye2025imgedit,zhang2023magicbrush}.

\begin{figure}
    \centering
    \includegraphics[width=\linewidth]{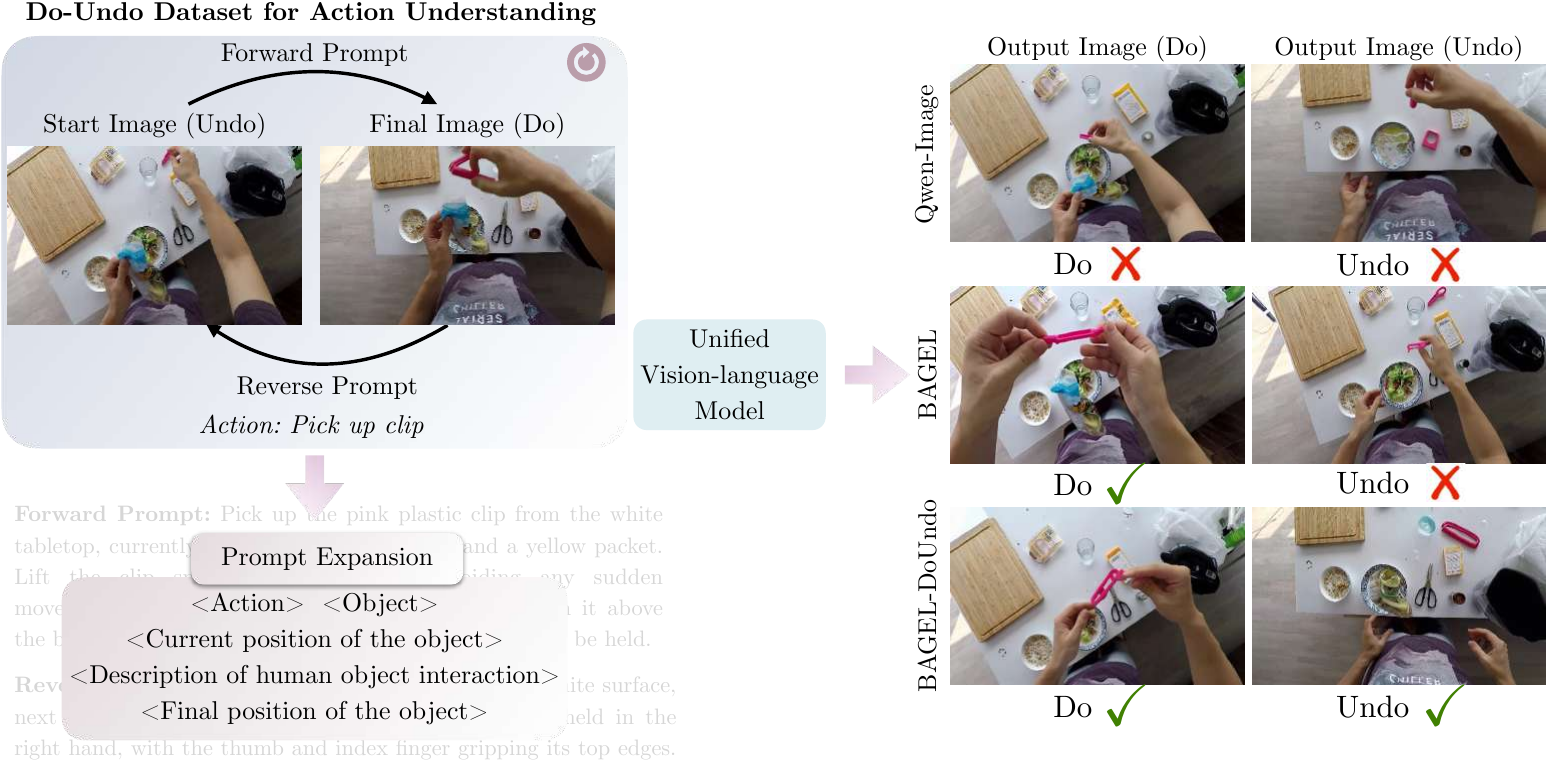}
    \caption{\textbf{Do-Undo task for action-conditioned image generation highlights a key limitation of current VLMs: their inability to reverse previously executed actions}. Models trained with \emph{Do-Undo} dataset show improved understanding of physical actions and their effects on scene dynamics.}
    \label{fig:teaser}
    \vspace{-20pt}
\end{figure}

For VLMs to be effective synthetic data generators in real-world applications such as in robotics and in embodied AI agents \cite{sang2023scene,yang2024physcene,lu2023synthetic,bhattacharyya2018long}, it is essential that they comprehend how physical actions transform the environment and generate images that plausibly reflect these changes. To make VLMs \emph{action aware} on classical mechanical manipulations, they should be able to generate the final state without observing a continuous sequence as in video models \cite{souvcek2024genhowto,trusca2024action}.
For example, given an image with an open refrigerator in a kitchen setting in \Cref{fig:teaser} and the action prompt ``pick up the clip", the model should be able to simulate a scene with a clip in hand without having to observe the entire sequence of lifting the clip. 
Furthermore, the image should preserve the dynamics and properties of the original scene.
For image generation models, this implies modeling the cause-and-effect relationships by observing the current image and the action in the form of a text or instruction prompt, and generating the final image revealing the state of the manipulated object and of the visual context.

Recent work on instruction-based image editing \cite{sheynin2024emu, hui2024hq} has primarily focused on the addition or removal of individual objects; or on maintaining physical properties such as lighting and reflections \cite{pu2025picabench, cai2025phys}.
However, these approaches overlook the state manipulation resulting from actions, where specific objects undergo transformation or state changes as a result of an action, while the rest of the scene remains unchanged—a property that is essential for synthesizing images reflecting realistic action-driven modifications.
Moreover, action-conditioned image editing models that generate the final state directly from the input image and instruction prompt do not account for consistency with the input image and require minimal camera movement to preserve background coherence \cite{souvcek2024genhowto}.

We identify a fundamental limitation in current image-generating VLMs: the inability to generate action-consistent images and to understand the relationship between actions and object states.
As illustrated in \cref{fig:teaser}, for the input image and the action prompts, even state-of-the-art models like Qwen-Image \cite{qwen2025qwen25technicalreport} and BAGEL \cite{deng2025bagel} struggle to generate physically consistent images. 
The models either hallucinate new objects or are unable to synthesize images conditioned on the performed action.

To evaluate and address this gap, we propose the Do-Undo task and benchmark, which challenges models to generate images that accurately reflect the outcome of a physical action, and then to reverse the action. 
We hypothesize that an action-aware image generation model that genuinely understands physical actions should be able to reverse an action that it has just performed and generate physically consistent images. 
Through comprehensive evaluation, we demonstrate that current state-of-the-art models struggle with this task, highlighting the need for new approaches to advance action-aware generative modeling.
Our reversible formulation and evaluation protocol assimilates dynamic scenes and camera movements, allowing models to generate diverse and plausible final images, provided they can return to the original state by undoing the current action.
By establishing the Do-Undo benchmark, we aim to set a new testbed for developing and evaluating VLMs capable of understanding and generating the physical world, thus advancing research in reliable embodied agents.

To summarize, our contributions are: 
\begin{enumerate*}[label = (\roman*)]
    \item We introduce a novel \emph{Do-Undo} task formulation with reversible, real‑world action understanding that requires models to generate the visual outcome of an action and then accurately invert it to reconstruct the original scene. This forward–reverse requirement explicitly tests whether models capture cause‑and‑effect dynamics rather than relying on superficial semantic cues.  
    \item We curate a large-scale dataset and benchmark of reversible actions with starting and final action states extracted from real-world videos in the Epic-Kitchens dataset \cite{damen2020epickitchensdatasetcollectionchallenges}. We design a specialized prompting strategy with forward and reverse action prompts to ensure physically consistent visual generation. Our benchmark accommodates dynamic scenes and camera movements and encourages models to generate diverse yet reversible images.
     \item We demonstrate that current state-of-the-art models struggle with the Do-Undo task by evaluating their performance on the Do-Undo benchmark, demonstrating a fundamental gap in current generative modeling—an inability to reason over actions and their consequences.
    \item We develop a baseline by training an image understanding and generation method, BAGEL \cite{deng2025bagel}, on our proposed dataset. Our results show that explicit supervision on reversible actions improves the fidelity and consistency of generated transformations, highlighting the benefits of Do‑Undo as a training signal for action-aware VLMs.
   
\end{enumerate*}

\section{Related Work}
\myparagraph{VLM-based image generation and editing.}
Unified vision-language models \cite{chen2025blip3ofamilyfullyopen,gemini-flash-2025,wu2025qwenimagetechnicalreport,deng2025bagel,batifol2025flux} for joint understanding and generation of images and text demonstrate impressive results in text-based image editing. 
BAGEL \cite{deng2025bagel} with its interleaved training strategy for understanding and generation can be applied to image editing tasks.
FluxKontext \cite{batifol2025flux} introduces a unified image generation and editing framework based on rectified flow matching. 
Qwen-Image \cite{wu2025qwenimagetechnicalreport} extends the Qwen-VL \cite{wang2024qwen2vlenhancingvisionlanguagemodels} understanding model to image generation in a multi-task training with Qwen-VL \cite{wang2024qwen2vlenhancingvisionlanguagemodels} for image understanding and a variational encoder (VAE) for image generation.
This enforces semantic coherence and high fidelity in image editing.
Generation chain of thought (GoT)~\cite{fang2025got} proposes a reasoning-guided paradigm for image generation and editing, incorporating both vision-text understanding and a semantic spatial module.
In addition to open models, proprietary models including Gemini \cite{gemini-flash-2025} and GPT-5 also provide image editing functionality.
We evaluate the VLMs, capable of image understanding and generation, for action awareness on our Do-Undo benchmark.

\myparagraph{Text-based image editing datasets.}
Text-based image editing datasets such as InstructPix2Pix \cite{brooks2023instructpix2pix}, EMU-Edit \cite{sheynin2024emu} and HQ-Edit \cite{hui2024hq} introduced synthetic datasets for instruction-based image editing.
InstructPix2Pix and EMU-Edit provide open-domain instructions on synthetic and real images respectively.
SEED-DataEdit \cite{ge2024seed} extends text-guided image editing to multi-turn scenarios. 
MagicBrush \cite{zhang2023magicbrush} provides an instruction-guided real image supporting single-turn, multi-turn, mask-provided, and mask-free editing.
However, these datasets lack action-guided editing instructions that reflect changes based on physical actions performed on objects. 
To evaluate instruction-based editing, Magicbrush provides a test set with and without masks as additional guidance for single and multi-turn editing. 
EMU-Edit extends MagicBrush with more challenging instructions, covering categories such as background and style manipulation, object removal and addition, texture changes, and global image modifications.
These benchmarks employ metrics such as CLIP~\cite{radford2021learning}, $\ell_1$, DINO~\cite{caron2021emerging} similarity, and human scores. We leverage the CLIP and DINO similarity scores to validate the semantic awareness of different models on our Do-Undo benchmark.

\myparagraph{Action-aware image editing.} GenHowTo \cite{souvcek2024genhowto}  and Aurora \cite{krojer2024learning} are the two action-centric editing datasets. 
GenHowTo samples frames from action-centric instructional videos with their captions starting from input image (which may or may not contain the target objects on which action is being performed) to an image showing action being performed and the final state. The dataset introduces new objects in the scene causing considerable drift from the input image limiting application to action-based image editing.
Aurora~\cite{krojer2024learning} covers a wide range of actions where the input and final state images share the same visual context, but often lack explicit causal clues, such as the presence of a person or hand manipulating objects. 
Our Do-Undo dataset addresses these limitations by providing high-quality, reversible action pairs with detailed context.


Aurora-bench further includes action-conditioned and reasoning-based editing instructions. The benchmark, in addition to the standard editing metrics, includes a score where the similarity between the input image and the two images generated with instructions that cause no change and considerable modification, respectively, is compared to measure the understanding of the instructional prompt by the editing model.
PICABench \cite{pu2025picabench}, introduces a physics-aware benchmark with physical effects such as optics, mechanics, and transitions, for example, in an image with ``a person riding a scooter." If the instruction is ``remove scooter", then the model should generate a physically plausible image with the person standing and not floating in the air.
In our benchmark, we study the action-following capabilities of VLM-based editing models based on the ability to reverse the performed action.

\begin{figure*}
     \centering
    \includegraphics[width=\linewidth]{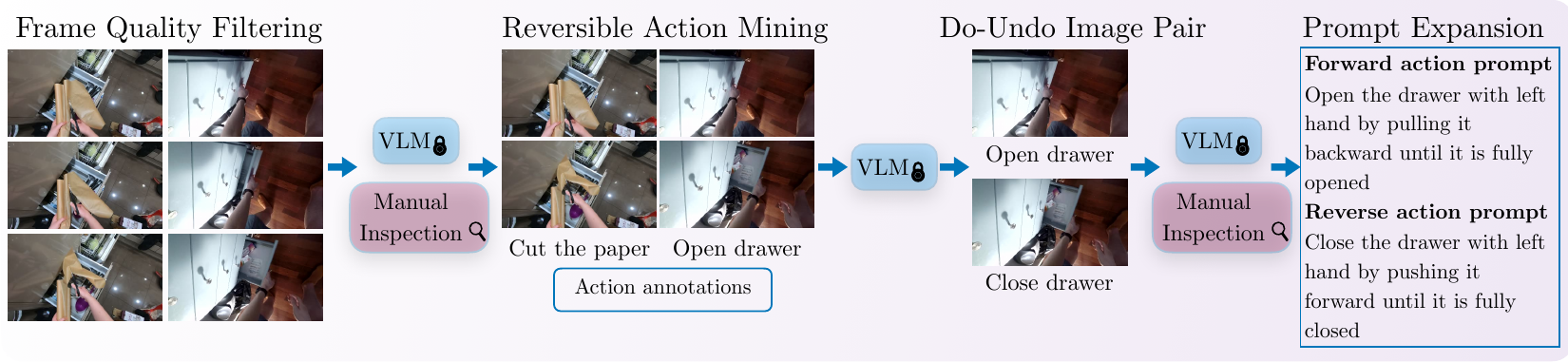}
    \caption{\textbf{Do-Undo data curation pipeline.} Starting with the EpicKitchens dataset \cite{damen2020epickitchensdatasetcollectionchallenges}, we select visually high quality samples which have reversible actions; the action annotations and the images are used to expand the prompts with visual context.}
    \label{fig:datacuration}
    \vspace{-10pt}
\end{figure*}
\section{Do-Undo Reversible Action-aware Task and Dataset}

In this section, we first motivate and describe our task, followed by the details of our Do-Undo dataset for training and the Do-Undo benchmark. 

\subsection{Do-Undo Task}
In this task, we investigate the capability of vision-language models designed for understanding and generation to synthesize images consistent with the action described by the input prompt. 
We consider an input image that contains an object on which an action is about to be performed, an agent (for example, a hand) performing the action, and the environment in which the interaction takes place.
If a VLM understands the current state of the input image, the action prompt, and the consequence of the action, it can synthesize a \emph{Do} or the \emph{forward} image, \ie the image after the action has been performed.
For this, VLM should
account for the visual content of the input image, including the state of the objects and the context; the action to be performed; the physics of action, object, and agent interaction; and generate a plausible image representing the visual state after the action is performed. 

A question that naturally emerges is that if the action is physically reversible in the real world, that is, one can obtain the original state by performing a complementary action, then a VLM should be able to reverse the action and generate the initial state (the \emph{Undo} image) given the corresponding reverse action prompt.
For instance, the action ``open the drawer'' can be reversed with ``close the drawer''; however, the action ``cut the paper'' is typically irreversible.
The ability to perform such reversible actions further instills action understanding in unified vision-language models.
To this end, we design a new \emph{Do-Undo} task by introducing a benchmark consisting of image and prompt pairs with reversible actions. 
This enables evaluating the unified vision-language models on their ability to model action-conditioned outcomes consistent across forward and reverse image generations.
Additionally, to show that intuitive tasks such as Do-Undo can induce implicit action-understanding, we provide training data with reversible action annotations.



\begin{figure*}[!t]
  \centering
  \begin{subfigure}[b]{0.45\textwidth}
    \includegraphics[width=\linewidth]{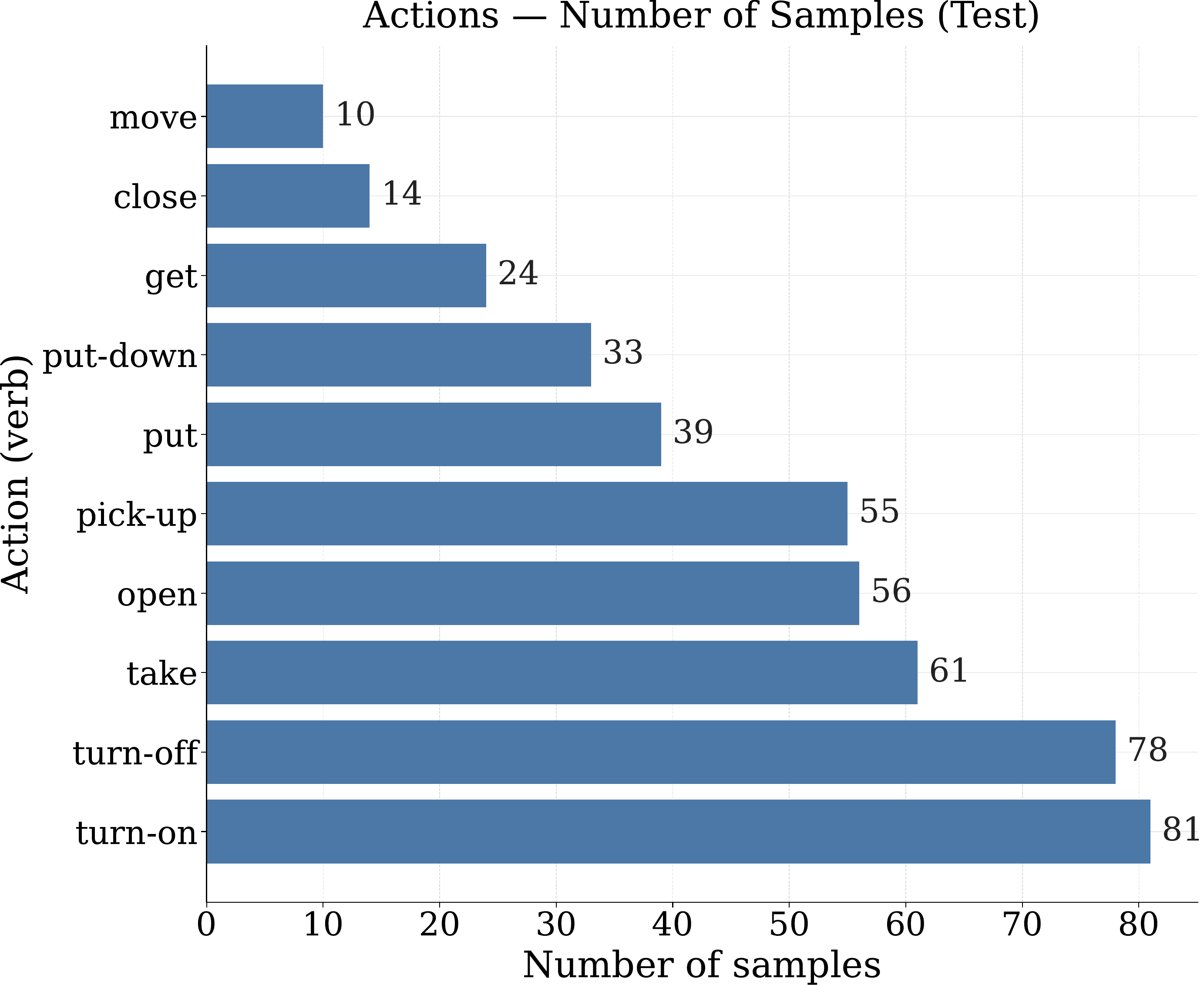}
    \caption{Number of samples and actions in the test set.}
    \label{fig: actionvssamplestest}
  \end{subfigure}\hfill
  \begin{subfigure}[b]{0.5\textwidth}
    \includegraphics[width=\linewidth]{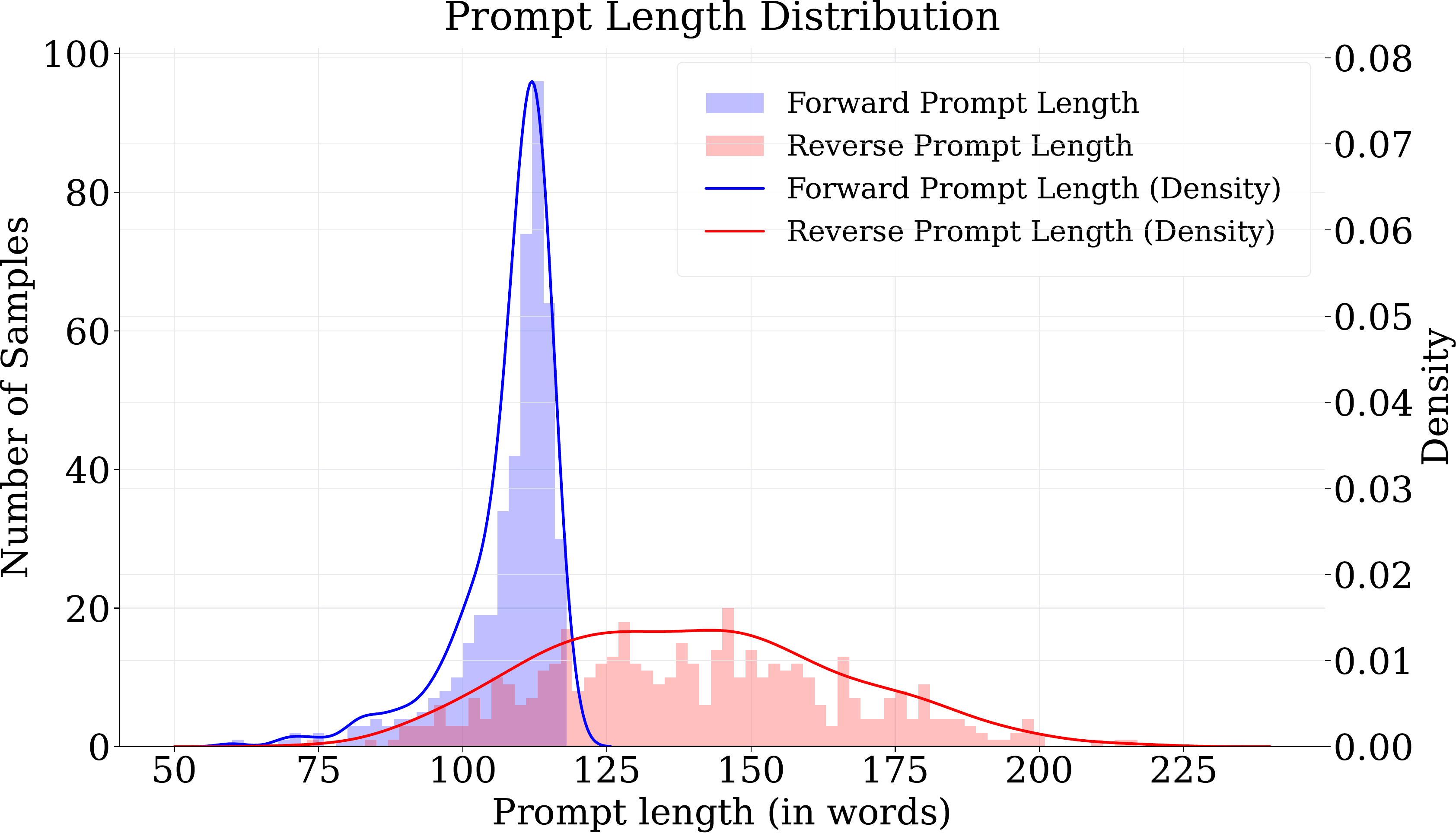}
    \caption{Prompt expansion statistics on the test set.}
    \label{fig:promptstatstest}
  \end{subfigure}
  \caption{\textbf{Dataset statistics of our Do-Undo test set.}  \textit{(left)} We show the distribution of actions in the test data. \textit{(right)} The test set includes prompts to guide the models for action-aware image generation.}
  \label{fig:datastatstest}
  \vspace{-10pt}
\end{figure*}

\subsection{The Do-Undo Dataset}
To support our \emph{Do-Undo} task for evaluating action-aware image generation, we construct \emph{a dataset centered on reversible actions in the real-world interactive environments, such as kitchens}.
By providing unified VLMs with scenarios where actions, their consequences, and their reverse counterparts are well-defined, we can probe whether these models truly understand how actions transform the world as reflected in images they generate.
Our datasets has the following key components:
\begin{enumerate*}[label = (\roman*)]
    \item \textit{Paired state transitions.} It consists of image pairs with start and end states, depicting the scene before and after an action has been performed, respectively.
    \item \textit{ Reversibility by design.} Every action is chosen to be physically reversible, ensuring that a return to the original state is not only feasible but also visually coherent and realistic.
    \item \textit{Embodied interactions.} The images include interaction between a human or an agent and the objects on which the action is applied. 
    \item  \textit{Action-conditioned prompts.} The start image is paired with a forward prompt, outlining the action and the object to be manipulated, along with a description of the environment. Analogously, the final state is paired with a reverse prompt with the reverse action and the descriptions of the object and environment. 
\end{enumerate*}
Formally, we collect a set of tuples $(\mathbf{I}_\text{o}, {P}_\text{F}, \mathbf{I}_\text{F},{P}_\text{R})$ with $\mathbf{I}_{\text{o}}$ being the input image, $\mathbf{I}_\text{F}$ is the image after the action has executed, ${P}_\text{F}$ is the forward action prompt, and ${P}_{\text{R}}$ is the reverse action prompt. In the following, we describe the data curation process of our dataset, outlined in \Cref{fig:datacuration}.

\myparagraph{Frame quality filtering and image pair acquisition.} 
To collect the image pairs with the start state image, $\mathbf{I}_\text{o}$, and the final state image $\mathbf{I}_\text{F}$, we rely on the Epic- Kitchens video dataset \cite{damen2020epickitchensdatasetcollectionchallenges}.
The  tasks in Epic-Kitchens are relevant to daily life and do not require specialized knowledge. These qualities make cooking a robust environment for our study.
Epic-Kitchens consists of 100 video episodes with subsequences comprising video frames from the start to the end of action. 
The videos feature humans performing tasks in a kitchen environment, recorded with an ego-centric camera set-up providing real-world, first-person perspective. 

To collect high-quality samples, we first exclude images with inadequate lighting or blur, where the visual content is difficult to interpret.
Following this, we identify suitable  start and end frames within a video sequence by employing Qwen2-VL-7B-Instruct \cite{wang2024qwen2vlenhancingvisionlanguagemodels} and
utilizing the action annotations provided in the Epic-Kitchens dataset for each sequence.
Starting from the start and end frames of a video sequence, the Qwen model checks for background and action consistency. 
Background consistency between the two frames is established based on minor camera movements and maintenance of the scene context, for example, the unchanged positions of the objects on which no action is being performed. 
Action consistency is confirmed by ensuring that the start frame is the state at the start of the action and that the final image reflects the scene state after the action has been performed. The Qwen model examines the state of the manipulated object in the final image as evidence for action completion.
Additionally, we use an action classifier \cite{zhao2023learning} to exclude frames with missing actions or target objects, yielding start and final images where the start and end of the action are clearly demonstrated.
Human annotators then perform a secondary verification step.

\myparagraph{Reversible actions.}
We first list a set of action vocabulary with their physically plausible reverse actions, including \emph{pick-up, put-down, put, open, grab, turn-off, turn-on, close, put-down, place, move,} and \emph{remove}.
It is worth noting that the action and its reverse can be in any order. That is, a ``turn-on'' action can happen before ``turn-off'' action or vice-versa. 
Moreover, different action descriptions can have the same inverse. In our case, ``grab'' and ``pick-up'' forward actions can be reversed with ``put'' or ``put-down''.
We consolidate image pairs with the action annotations based on this vocabulary.

\myparagraph{Prompt expansion for action-conditioned prompts.}
Since EPIC‑Kitchens \cite{damen2020epickitchensdatasetcollectionchallenges} provides short action narrations with an average length of only three words, we introduce a prompt expansion strategy to make the dataset suitable for instruction‑following in vision–language models. 
With this, we enrich the action prompts with additional visual and contextual information based on the input sequence.
To construct each forward‑action prompt $P_{\text{F}}$, we provide Qwen3‑VL \cite{wang2024qwen2vlenhancingvisionlanguagemodels} with five temporally sampled frames from the action sequence along with the original EPIC‑Kitchens action annotation. 
The model is instructed to generate a prompt that preserves the \texttt{<action, object>} structure of the annotation while expanding it with richer contextual information. 
Specifically, the expanded prompt describes the manipulated object using attributes such as material, color, and its spatial position before the action.
The prompt includes the details of a person's hand (one hand, two hands, posture, left or right hand) and the
spatial relationship between the object and the human.
In addition to this, the prompt provides the desired state or location of the object after the action has been performed.
Thus, each prompt provides a detailed and semantically grounded description of both the action and its intended outcome.

Analogously, we provide the frames in reverse order and provide the same instructions to create the reverse prompt  $P_\text{R}$ to undo the action.
These action prompts guide the VLM for precise image editing while accounting for the variations in camera movements or the background mismatch between the start and the end images. 
The complete instruction prompt is provided in the appendix.

\myparagraph{Dataset statistics.}
After curating the Epic-Kitchens dataset with reversible actions, we evaluate the performance of different models for action-grounded understanding and generation on our \emph{Do-Undo benchmark}. 
To ensure fairness in the benchmark, the video sequences used to construct the test data are sourced from the test portion of the Epic-Kitchens dataset, with no overlap with those of the training set.
The test data is balanced across actions with a total of 451 samples. As shown in \Cref{fig: actionvssamplestest}, the ten action classes are well represented in our test data.
Noting that vision-language models are sensitive to the prompt length, we provide long prompts designed with our prompt expansion strategy above, with an average prompt length of approximately 120 words \Cref{fig:promptstatstest}.

We obtain 22,529 samples in the training set, including both the forward and reverse action pairs with a total of 45,058 annotations. 
In \Cref{fig:datastatstraining}, we analyze the samples from the training data. 
The joint vocabulary of action (verb) and object (nouns) \texttt{<action,object>} pairs provides sufficient sample diversity.
As shown in \Cref{fig:nounvsaction}, even though pick-up is the most frequent action type, it is accompanied by a diverse set of object or noun types.
This balances out more pronounced actions, such as pick-up, with almost 26\% of the action annotations in training data (\Cref{fig: samplesvsactions}).

\subsection{Fine-tuning on Do-Undo}
\label{sec:method}
To show the advantages of training with our \emph{Do-Undo} paradigm, we assume a vision-language model (VLM) with the capability to generate both images and text in an interleaved setup; $\mathcal{E}_{\theta}$ parameterized by $\theta$ that takes as input an image and prompt to generate images.
By training a VLM on our Do-Undo training set, we aim to induce image understanding and generation grounded in actions by enforcing consistency between the synthesized images for the forward and the reverse actions. 
In our work, we consider BAGEL \footnote{\scriptsize Only large-scale unified multimodal model with available training code (under Apache 2.0 license at the time of submission).} \cite{deng2025bagel} as the underlying baseline VLM. 

Our training data consists of tuples $(\mathbf{I}_\text{o}, {P}_\text{F}, \mathbf{I}_\text{F},{P}_\text{R})$.
Let $\mathbf{I}_{\text{o}} \in \mathbb{R}^{H\times W \times3}$  be an input image and $P_{\text{F}}$ be a reversible action prompt describing a physically meaningful manipulation of objects in $\mathbf{I}_{\text{o}}$ (e.g., “open the drawer with left hand by pulling it backward until it is fully opened”). 
We first encode $P_\text{F}$ and $\mathbf{I}_{\text{o}}$ using a text tokenizer and a ViT, respectively.
The combined features form the context for the subsequent generation of frame ${\mathbf{I}}_{\text{F}}$. 
Rectified flow matching \cite{DBLP:conf/iclr/LiuG023} is employed as a conditional image generation model minimizing the mean squared error (MSE) for noisy encoding of ${\mathbf{I}}_{\text{F}}$ with a VAE encoder yielding $\hat{{\mathbf{I}}}_{\text{F}}$.
During training for the reverse direction, we encode the reverse action prompt $P_\text{R}$ and the groundtruth image $\mathbf{I}_{\text{F}}$ into the VLM which serve as the conditioning or context for generating the reverse image $\hat{\mathbf{I}}_\text{R}$. 
Notably, the generated undo image should be the same as the input image $\mathbf{I}_{\text{o}}$. 
Therefore, for generating the reverse image with rectified flow, we minimize the mean-squared error with respect to the noisy latent from $\mathbf{I}_{\text{o}}$ to get 
$\hat{\mathbf{I}}_\text{R}$.
We follow the finetuning strategy of the original model \cite{deng2025bagel} for multimodal understanding and generation tasks: text-to-image generation on the image-text-pair set; \emph{interleaved training on our Do-Undo dataset} and multimodal understanding on the instruction finetuning set from BAGEL \cite{deng2025bagel}.  The model is trained on the mean-squared error from rectified flow matching and the cross-entropy loss for next-token prediction. 

\section{Experiments}

\label{sec:experiments}
To investigate the performance of unified VLMs for action-aware generation, we show zero-shot performance of
Qwen-Image \cite{wu2025qwenimagetechnicalreport} and Flux-Kontext \cite{batifol2025flux} on our proposed \emph{Do-Undo} benchmark. 
We further compare BAGEL \cite{deng2025bagel} against BAGEL-DoUndo, i.e., BAGEL fine-tuned on our training set.
\begin{figure*}[t]
    \centering
    \includegraphics[width=\linewidth]{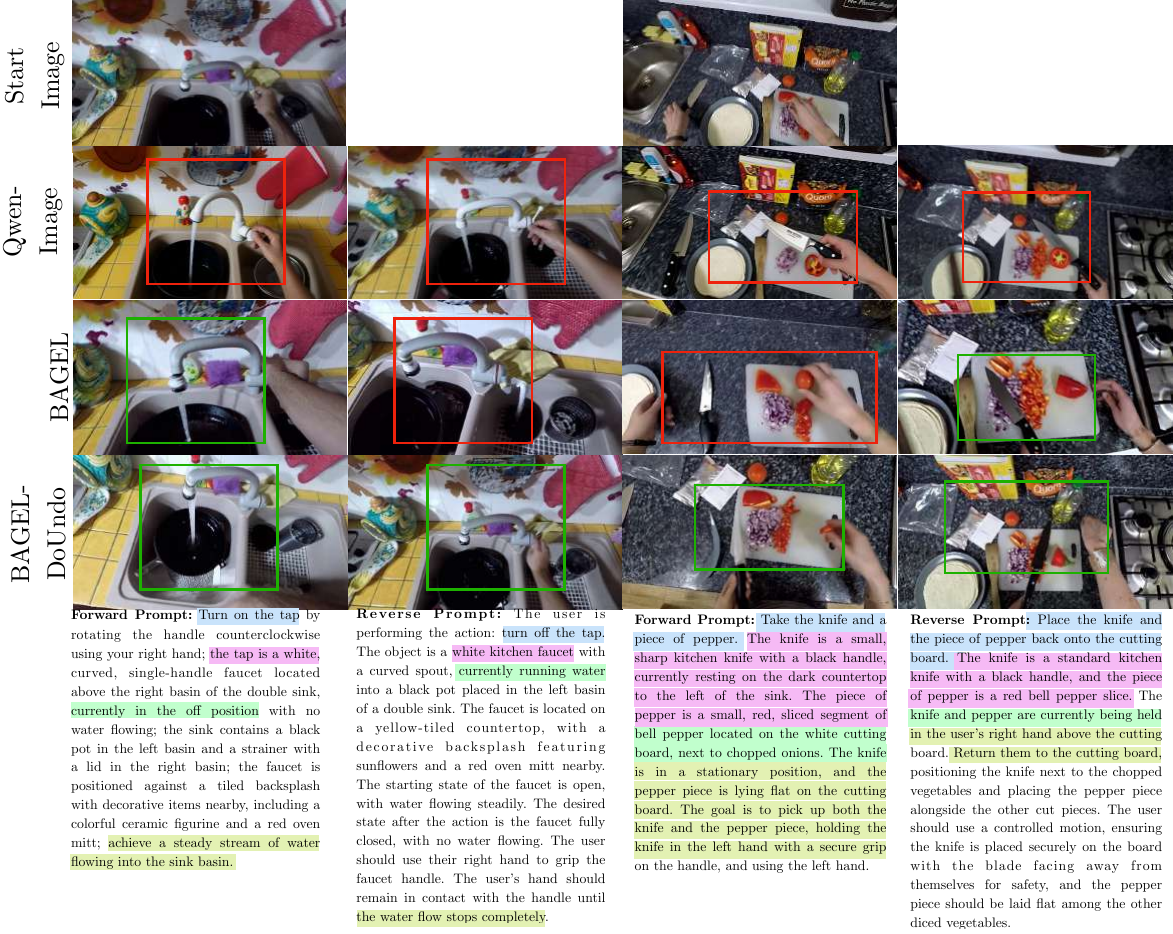}
    \caption{\textbf{Qualitative results.} Qualitative comparison of Qwen Image \cite{wu2025qwenimagetechnicalreport}, BAGEL \cite{deng2025bagel} and BAGEL-DoUndo (Ours) on our benchmark. Qwen-Image does not preserve
the object semantics and cannot faithfully perform the reverse action. BAGEL struggles with modeling object states after the action has been performed. BAGEL-DoUndo approach consistently generates images that adhere to the semantics of the input image, including human object interaction.}
    \label{fig:qualitative}
    \vspace{-10pt}
\end{figure*}
\begin{table*}[t]
\centering
\caption{\textbf{Zero-shot evaluation}. Different VLMs struggle to perform on our Do-Undo benchmark. High semantic fidelity does not inherently translate to superior action understanding. }
\label{tab:zeroshotlong}
\footnotesize
\begin{tabularx}{\linewidth}{@{}Xccc|cccccc@{}}
\toprule
{Method} & DINO-F &  DINO-R  & CLIP & A-F  & A-R & N-F & N-R & EPE $\downarrow$ &  EPE-R $\downarrow$ \\
\midrule
Qwen-Image \cite{qwen2025qwen25technicalreport}   & 0.817 & 0.815 &  0.258  &  52.33 & 29.71 & 61.20 &  52.77 & 89.23 & 80.86 \\
Bagel \cite{deng2025bagel}  & 0.793 & 0.796  &  0.262 &  57.87 & 33.48 & 55.65  & 50.55  & 121.0 & 94.07 \\
Flux Kontext \cite{batifol2025flux}  & 0.750 & 0.746   & 0.240 & 52.23 & 30.12 & 53.23 & 48.18  & 111.2 & 95.87 \\
\bottomrule
\end{tabularx}
\vspace{-10pt}
\end{table*}
\begin{table*}[t]
\caption{\textbf{Quantitative results.} Evaluation on  BAGEL and the variants with Do-Undo training set shows that BAGEL-UnDo has high accuracy for action understanding compared to the baseline.}
\centering
\footnotesize
\begin{tabularx}{\linewidth}{@{}Xccc|cccccc@{}}
\toprule
&  \multicolumn{3}{c}{Semantic Awareness} & \multicolumn{6}{c}{Action Understanding}  \\
\cmidrule(lr){2-4} \cmidrule(lr){5-10} 
Method &  DINO-F & DINO-R  & CLIP & A-F  & A-R & N-F & N-R & EPE-F$\downarrow$ & EPE-R$\downarrow$  \\
\midrule
BAGEL \cite{deng2025bagel}  & 0.796  & 0.793 &  \bfseries 0.262 & 57.87 & 33.48 & 55.65 & \bfseries 50.55 & 121.0 & 94.07  \\
BAGEL-Do(SP)  & 0.818 & 0.819 & 0.254  & 55.65& 34.81& 54.55& 47.23& 118.8 & 93.27 \\
BAGEL-Do  & 0.821 & 0.816 & 0.250    & 55.92 & 34.60 & 56.87 & 46.21 & 124.5& 93.70\\
BAGEL-DoUndo  & \bfseries 0.836 & \bfseries 0.832  &   0.251    & \bfseries 58.77 & \bfseries 36.26 & \bfseries 58.53& 50.47 & \bfseries 118.4  & \bfseries 90.88\\
\bottomrule
\addlinespace
\toprule
BAGEL(multiturn)   &  0.830  & 0.850  & 0.251 & 54.55   & 35.22  & 54.77 & 46.76 & \bfseries  99.55 & \bfseries  66.16\\
BAGEL-DoUndo(multiturn)& \bfseries 0.831 & \bfseries   0.872  &  0.251 & \bfseries  56.76  &  \bfseries 37.92 &  \bfseries  57.65   &  \bfseries 48.78 &  116.12  &  74.30 \\
\bottomrule
\end{tabularx}
\vspace{-10pt}
\label{tab:main_results}
\end{table*}

\myparagraph{Evaluation metrics.} We validate the performance on a diverse set of metrics.
The metrics are divided into two categories that evaluate semantic awareness and action understanding. 
Specifically, the metrics evaluating semantic awareness are:
\begin{enumerate*}[label=(\roman*)]
\item DINO-F measures the similarity between the generated forward image $\hat{\mathbf{I}}_{\text{F}}$ and the ground-truth ${\mathbf{I}}_{\text{F}}$.
\item Similarly, DINO-R measures the image similarity between the reverse image $\hat{\mathbf{I}}_{\text{R}}$ and the original input image $\mathbf{I}_{\text{o}}$. It evaluates the ability of a model to generate the semantic content consistent with the original state. 
\item To account for diversity in generated images, we measure the CLIP similarity of the generated image $\hat{\mathbf{I}}_{\text{F}}$  with the caption of the ground-truth image. 
\end{enumerate*}
To evaluate action understanding in vision-language models, we include a diverse set of metrics:
\begin{enumerate*}[label=(\roman*)]
\item We build an action classifier by leveraging the action recognition capability of LaViLa \cite{zhao2023learning}. 
We finetune the model on our Do-Undo training and test set to achieve oracle performance. Following this, we compute the action accuracy of the forward image (A-F) and that of the reverse image (A-R). 
Furthermore, we include the accuracy of the generated objects (nouns) given by N-F and N-R for the forward and reverse images, respectively.
\item We include the optical flow-based error (EPE-F) using RAFT \cite{teed2020raft}. 
To quantify the error, we calculate the mean‑squared difference between the forward optical flow estimated from the start to the forward image and the ground-truth flow between the start and the ground-truth forward image.
\item Additionally, we include optical flow error between the reverse image and the ground-truth image (EPE-R).  
\end{enumerate*}

\begin{figure*}[t]
    \centering
\includegraphics[width=0.8\linewidth]{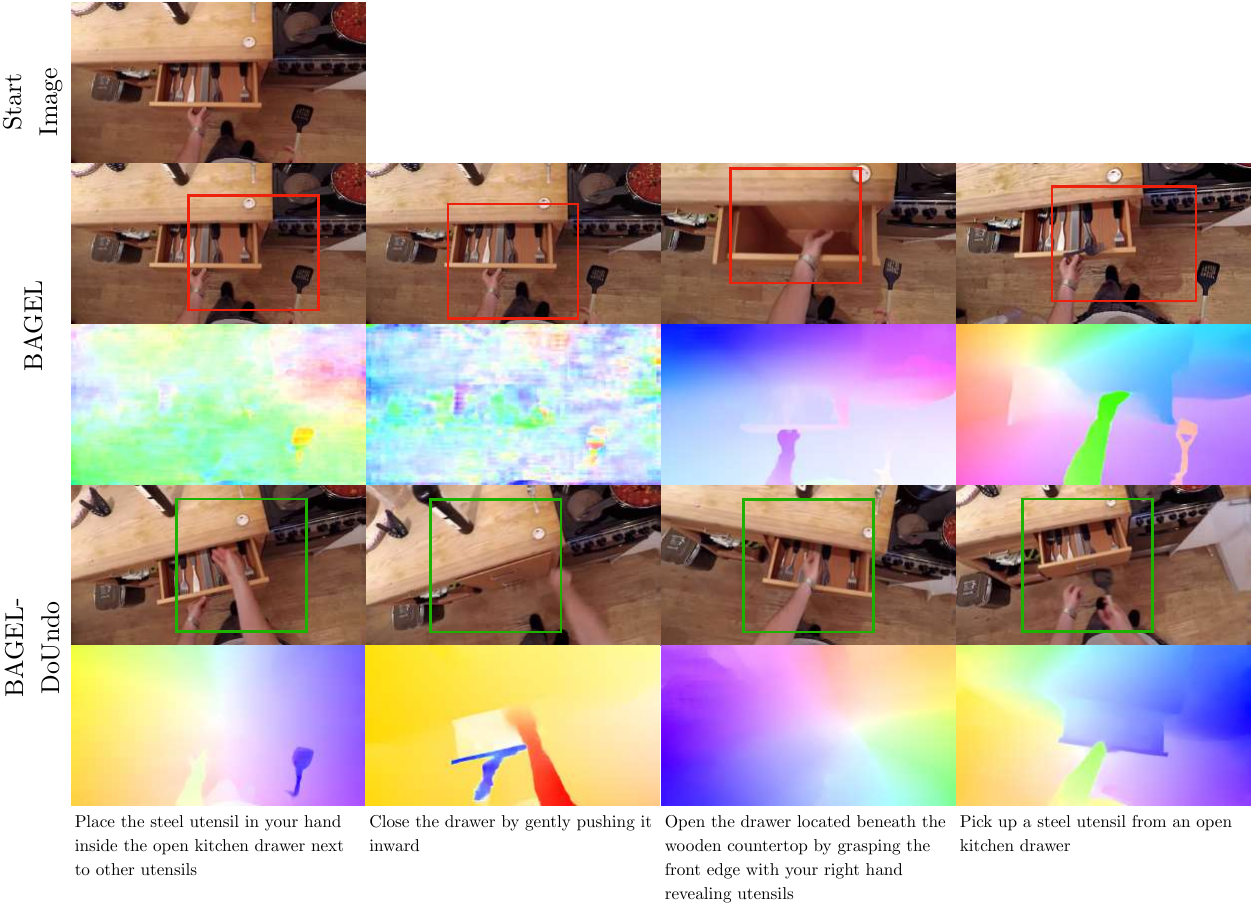}
    \caption{\textbf{Multi-turn multi-action-conditioned generation.} Each generated image is guided by the previously generated images and prompts that serve as context. The optical flow maps show that BAGEL-DoUndo accurately manipulates the target objects while preserving the visual context.}
    \label{fig:multiturn}
    \vspace{-20pt}
\end{figure*}

\subsection{Quantitative Results} 

\myparagraph{Zero-shot evaluation.}
\Cref{tab:zeroshotlong} shows the results of zero-shot evaluation on our Do-Undo benchmark on different state-of-the-art unified generation models: Qwen-Image \cite{qwen2025qwen25technicalreport}, BAGEL \cite{deng2025bagel}, and Flux-Kontext \cite{batifol2025flux}.
These models have a dual training pipeline for image understanding and generation, making them an ideal test-bed for evaluating the prompt-based action understanding. 
With reversibility, since only the object and the agent acting must be manipulated based on the prompt, Do-Undo is well-suited to evaluate action-conditioned relationships, while other factors remain unchanged.
We observe that while these models achieve high semantic awareness by generating images with high visual fidelity, they exhibit poor performance on evaluation metrics for action understanding, A-F and A-R. 
This gap is most evident in Qwen-Image, which maintains strong DINO-F and DINO-R scores ($\approx 0.81$), yet lags in action accuracy with $52.3\%$ and $29.7\%$ A-F and A-R, respectively.
Evidently, from the semantic and action understanding scores of BAGEL and Qwen-Image, high semantic awareness does not correlate with high action understanding, with BAGEL performing better than Qwen-Image on action understanding metrics. \emph{These findings highlight a critical limitation in current unified models: the ability to generate high-fidelity images does not guarantee the ability to model the state changes induced by specific actions.}

\myparagraph{Finetuning with Do-Undo.} \Cref{tab:main_results} shows the quantitative evaluation of the BAGEL baseline against the BAGEL-DoUndo approach on our {Do-Undo benchmark}. 
As demonstrated, our approach outperforms BAGEL across semantic awareness and action understanding metrics, with A-F and A-R of 58.77\%  and 36.26\% compared to 57.8\% and 33.48\%, respectively. 
Baseline BAGEL struggles to generate semantically consistent images, as reflected by lower DINO similarity scores.
These results highlight the benefits of incorporating reversible-action understanding and generation through the Do-Undo task for action awareness in VLMs (\cf~\Cref{fig:statescore}). 

\myparagraph{Effect of prompt expansion and reverse image pairs.} To verify the contributions of the different components of the Do-Undo paradigm, specifically, prompt expansion and the reverse image pairs, we derive the following variants (\cf~\cref{tab:main_results}~rows ~2 \& 3).
BAGEL-Do(SP) is trained only on the forward (Do) images with short narrations, <\texttt{action, object}>. 
BAGEL-Do, in contrast, is trained on the forward images with long prompts generated via prompt expansion. 
While BAGEL-Do(SP) and BAGEL-Do improve the semantic alignment relative to BAGEL, the performance on action understanding declines. 
This shows that training with only the forward images and prompts fails to induce action-conditioned understanding and generation in VLMs.
The significant performance gain of BAGEL-DoUndo over BAGEL-Do, validates the benefit of reverse image pairs.

\myparagraph{Multi-turn and multi-action evaluation.} Furthermore, we extend the DoUndo task to a multi-turn setup where first the forward image is generated conditioned on the start image and the forward prompt (\cf~\cref{tab:main_results}~rows ~5 \& 6). 
To generate the reverse image, the reverse prompt, the generated forward image, the forward prompt, and the start image are provided as context. 
BAGEL and BAGEL-DoUndo have similar performance on the semantic awareness, however, BAGEL-DoUndo shows better action understanding in terms of the action accuracy scores. 
The low EPE error of BAGEL(multi-turn) results from no camera movement or when no action is performed (\cf~Appendix~\ref{sec:supp:eval}).
\begin{wraptable}[6]{r}{0.45\linewidth}   
\vspace{-13pt}
\centering
\scriptsize
\caption{ \small
User study for semantic awareness and action understanding.
}
\begin{tabularx}{\linewidth}{@{}X c @{}}
\toprule
Method & Preference (\%) \\
\midrule
BAGEL-DoUndo (Ours) & \textbf{66.7} \\
BAGEL & 33.3 \\
\bottomrule
\end{tabularx}
\label{fig:user_study}
\end{wraptable}
Thus, we attribute performance gains in action understanding and action-aware image generation to our unique task formulation and training on our Do-Undo dataset, reflected in BAGEL-DoUndo.
The user study (\cref{fig:user_study}) further validates the performance gains of our approach, where BAGEL-DoUndo is preferred 66.7\%  on average compared to BAGEL with 33.3\% preference score (\cf~Appendix \ref{sec:Userstudy}) for details.

\subsection{Qualitative Results} 
In \Cref{fig:qualitative} we present the qualitative results of unified VLMs Qwen-Image\cite{qwen2025qwen25technicalreport}, BAGEL, and our BAGEL-DoUndo on the Do-Undo benchmark.
We notice that Qwen-Image (row~2) does not preserve the object semantics and generates textureless images.
Moreover, the model does not adhere to the action prompts and cannot faithfully perform the reverse action. 
BAGEL (row~3), on the other hand, generates textured images reflecting the frequency details of the start state image.
However, the model struggles with modeling object states after the action has been performed. 
For example, the water is flowing from the knob of the tap for the action prompt ``turn off the top''  (\cf.~\cref{fig:qualitative} row~3, col.~2).
Similarly, it fails to generate the image with a person holding a knife and pepper in col.~3.
Here, our BAGEL-DoUndo approach consistently generates images that adhere to the semantics of the input image, including human object interaction. 

\myparagraph{Multi-turn and multi-action generation.} \Cref{fig:multiturn} offers additional insights into action understanding, comparing BAGEL and our BAGEL-DoUndo approach in a multi-turn generation set-up through optical flow visualizations.
Starting from an initial image, each model is prompted to generate a forward state, after which it is instructed to perform the next action on its own generated output. This sequence is repeated for four steps.
Our approach not only generates the correct states but also maintains the background consistency, as shown in the optical flow maps. 
In the first column, the optical flow map shows that our generated image manipulates the target object (steel utensil) only, whereas BAGEL shows wider variations in the flow map.
Across the four images, BAGEL-DoUndo exhibits stronger semantic adherence and clearer action understanding.

\section{Conclusion}
We introduced Do-Undo, a new task and benchmark to assess the limitations of VLMs in understanding and generating physically plausible images based on real-world actions. 
Do-Undo emphasizes cause-effect reasoning for generating synthetic data by requiring the model to generate the forward action and then reversing it to go back to the original state.
Through our extensive experiments, we demonstrated that even the best-performing models struggle with feasible reversible actions and often hallucinate new objects or fail to maintain scene consistency. 
We believe that our new task and benchmark serve as an important testbed for the development of physics-aware generative models.\\
\textbf{Limitations and future work.} Our work builds upon the Epic-Kitchens dataset to enforce a controlled, yet real-world setting. The benchmark assumes that a wide range of general-purpose VLMs share the inductive biases from the dataset; without requiring specialized knowledge, for example, from a robotics dataset \cite{DBLP:conf/rss/KhazatskyP0BDKN24}. In future work, we aim to extend the benchmark to these specialized embodied environments. 
Furthermore, developing benchmarks to support action understanding and causal relationships through intuitive physics is a promising direction for future work.\\
\textbf{Broader impact.} This research aims to advance physical understanding in world models with applications in embodied AI. The Undo component also serves as an action interpretability tool to identify action understanding rather than prompt-image correlation. This work builds on VLMs, which are susceptible to biases and harmful content generation. Training and evaluation of large-scale models come with a high environmental cost.

{
    \small
    \bibliographystyle{plain}
    \bibliography{paper-template-Latest/main}
}

\clearpage
\section*{Appendix}
\appendix
The appendix is organized as follows:

\begin{itemize}
\item \Cref{sec:supp:datacleaning} provides details on data quality control.
    \item \Cref{sec:supp:promptexpansion} details the prompt expansion strategy and provides the prompt used to construct the benchmark and dataset.
    Additionally, we provide an empirical justification for using long prompts in comparison to short prompts.
    \item \Cref{sec:supp:eval} provides a discussion of the action accuracy metric and the limitations of EPE as a stand-alone evaluation metric. 
    \item \Cref{sec:Userstudy} provides additional details on the user study.
    \item \Cref{sec:supp:ood} extends qualitative evaluation to the in the wild setting where we show the generalization abilities of model trained on our training set to perform general actions in diverse environments.
    \item \Cref{sec:supp:additionalqualitatives} provides additional results with multiturn evaluation.
\end{itemize}

\begin{figure*}[!h]
  \centering
  \begin{subfigure}[!t]{0.5\textwidth}
    \centering
    \includegraphics[width=0.9\linewidth]{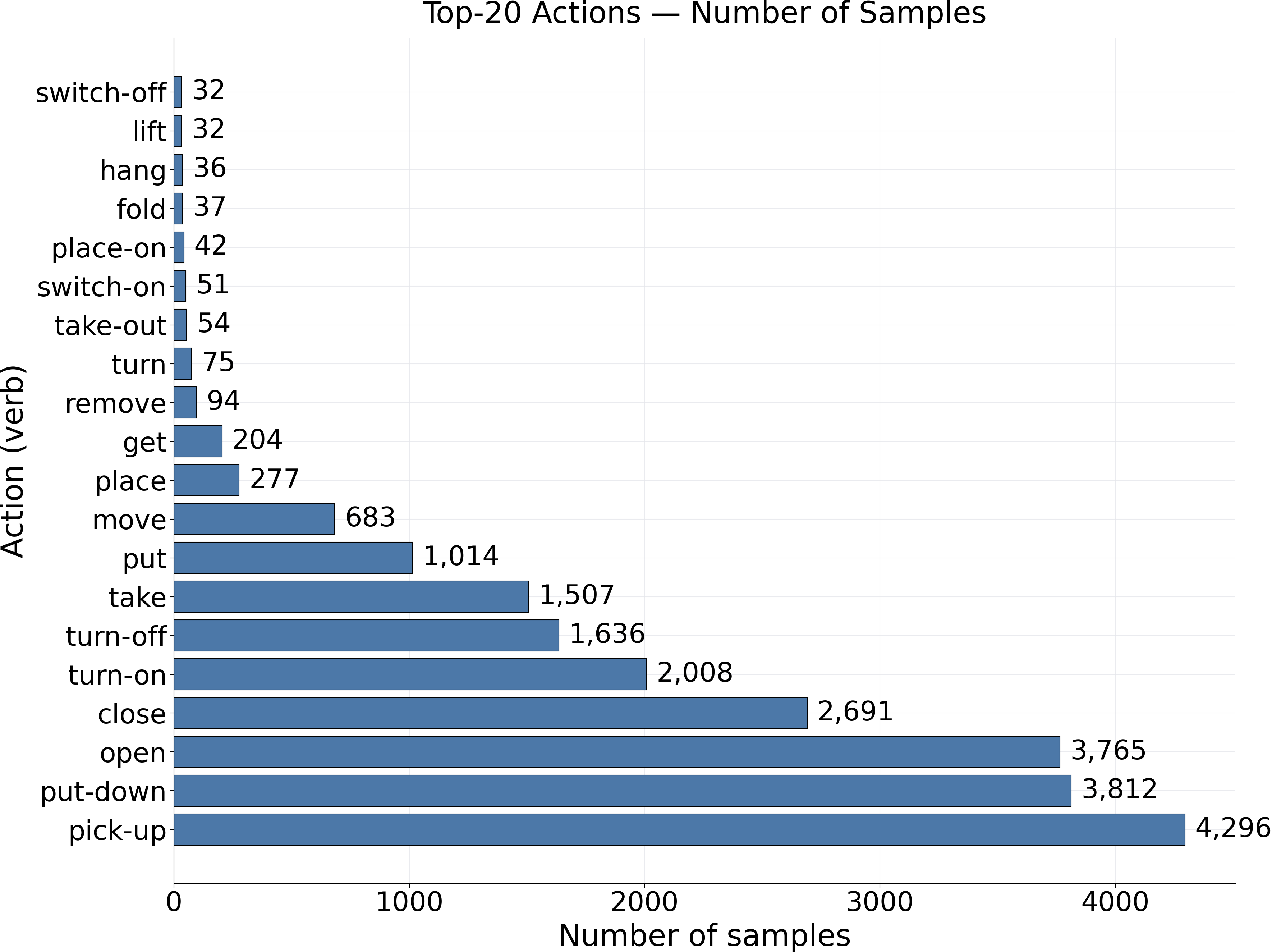}
    \caption{Number of samples for top-20 reversible actions in the training set.}
    \label{fig: samplesvsactions}
  \end{subfigure}\hfill
  \begin{subfigure}[!t]{0.5\textwidth}
    \centering
    \includegraphics[width=0.9\linewidth]{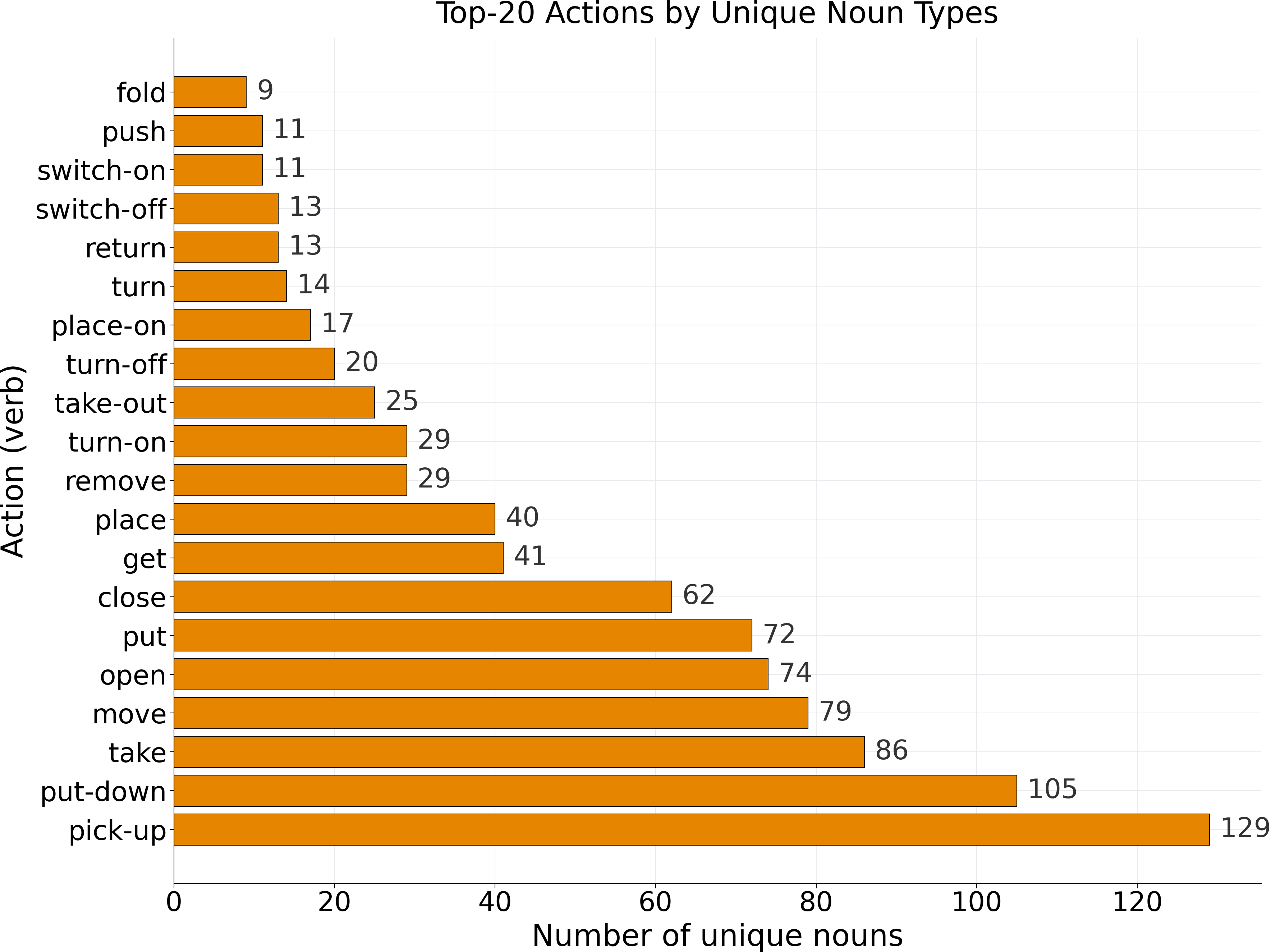}
    \caption{Number of unique nouns or objects for each action in the training set.}
    \label{fig:nounvsaction}
  \end{subfigure}
  \caption{\textbf{Dataset statistics of our Do-Undo training set.} We generate data by mining reversible tasks from the Epic-Kitchens \cite{damen2020epickitchensdatasetcollectionchallenges} dataset. \textit{(left)} We show the distribution of top-20 actions in the training data. \textit{(right)} We analyze the diversity of unique objects and actions.}
  \label{fig:datastatstraining}
\end{figure*}

\section{Data Cleaning and Quality Assurance Pipeline}
\label{sec:supp:datacleaning}
The dataset underwent a three-stage verification process to ensure temporal alignment, physical consistency, and high-fidelity grounding.
\begin{itemize}
    \item \textit{Temporal alignment and action classification}. To ensure frames with the start and end of the action, we employed a pre-trained action classifier on all samples.
    This step verified that the "Start" frame correctly depicts the initial state of the action and the "End" frame captures the completed state.
    \item \textit{Automated verification with Qwen-VL.} We employed Qwen-VL on the aligned frames to evaluate consistency between the prompts and the corresponding frames. 
    The model provides overall confidence scores: a quantitative measure of the alignment between the image pair and the action description; and validates if state change described in the prompt is visually reflected in the transition between frames.
    \item \textit{Manual verification.} Samples flagged with "Low" or "Moderate" confidence by the automated pipeline (approximately 265 frames) were diverted for manual review.  We reviewed these specific cases to make a final "Keep" or "Filter" decision, ensuring that subtle physical nuances or complex background interactions were handled correctly.
\end{itemize}
\section{Prompt Expansion}
\label{sec:supp:promptexpansion}

In the following, we provide the instruction provided to \texttt{Qwen3-VL-30b} to obtain the action-grounded prompts to guide an image-editing model for action-guided image synthesis. To generate the prompt, we provide as input the start state image and the end state image in addition to the narration of actions (action text) such as ``open door" in the Epic-Kitchens dataset. Notably, for the undo prompt generation, we reverse the order of the start and end state images.

The prompt contains the description of the action, the object on which the action is being performed,  and its description. 
Additionally, we provide the starting location described by the semantics of the real world, as well as the desired end location of the object after the action is performed. 
The prompts also provide instructions on how the user must interact with the object to perform the desired action.

\myparagraph{Discussion on Do-Undo benchmark.} 
\begin{table}[h]
\centering
\scriptsize
\vspace{-10pt}
\caption{Evaluation with short prompts.}\begin{tabularx}{\linewidth}{@{}Xcccc@{}}
\toprule
Method &  DINO &  CLIP & A-F & N-F \\
\midrule
BAGEL & 0.81 & 0.25 &  45.9 & 53.22\\
BAGEL-Do (SP)  & 0.84 & 0.24 & 53.22 & 52.33 \\
BAGEL-Do  & \bfseries 0.85 & 0.24 & \bfseries  54.55 & 50.33 \\
\bottomrule
\end{tabularx}
\label{tab:ablation_shortprompts}
\end{table}
To ensure that our benchmark supports action understanding, we conduct a study where we evaluate BAGEL, BAGEL-Do(SP), and BAGEL-Do on short prompts containing only \texttt{<action, object>} pairs in \Cref{tab:ablation_shortprompts} .
Even though short prompts can be used to synthesize semantically relevant images, they tend to yield low action accuracy. 
This supports the importance of prompt expansion within our benchmark to guide VLMs toward action-aware image generation.

\begin{tcolorbox}[
    colback=blue!5!white, 
    colframe=blue!75!black, 
    title=\textbf{Instruction Prompt for Prompt Expansion},
    fonttitle=\bfseries,
    coltitle=black, 
    colbacktitle=blue!20!white, 
    sharp corners,
    boxrule=0.5pt
]
    The user is performing the action: \texttt{action text}. \\
        Looking at this {\texttt{end} if \texttt{isreverse} else \texttt{start} frame, generate an expanded {direction} prompt.}  \\
        Structure the prompt with the action to be performed, object, and description of object, \\ starting, location or state of the object, and the location or state to achieve after the action. \\
        Also, add information on how the user should perform the action, such as object handling and hand interaction. \\
        Provide ONLY the expanded prompt string.
\end{tcolorbox}

\section{Evaluation Metrics and Computational Requirements}
\label{sec:supp:eval}
\begin{figure}[t]
    \centering
    \includegraphics[width=\linewidth]{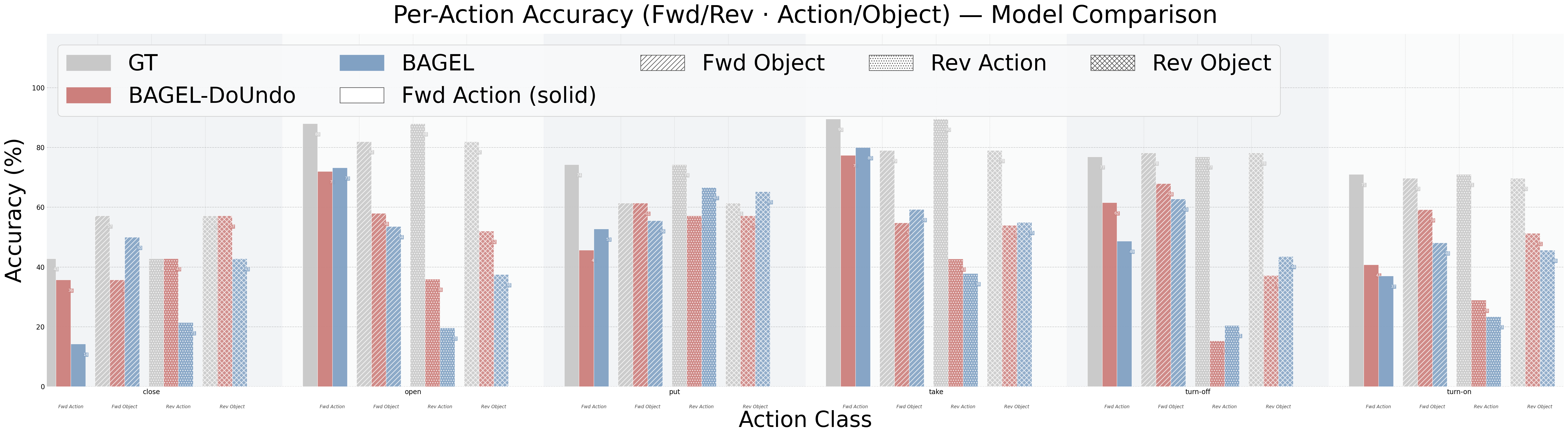}
    \caption{Action accuracy for different actions for the ground-truth, images synthesized by BAGEL and images synthesized with BAGEL-DoUndo on the BAGEL-DoUndo benchmark.}
    \label{fig:peractionaccuracy}
\end{figure}

In \cref{fig:peractionaccuracy}, we show the accuracy scores for each of the action classes in our Do-Undo benchmark for the baseline BAGEL \cite{deng2025bagel} and BAGEL-DoUndo.
The action accuracy verifies if the correct action is being performed, given the start state and the end state image.
As outlined in the main paper, we use LaViLa \cite{zhao2023learning} and finetune it on the Epic-Kitchens training set. 
Following this, we obtain an overall upper-bound action accuracy of 78.27\%  and noun accuracy of 72.51\% on the ground-truth images in our DoUndo benchmark. 
We observe that BAGEL-Do-Undo yields an action accuracy score of 58\% on average.
The gap between the ground-truth and the BAGEL-DoUndo approach highlights the complexity of the task and the benchmark for action-conditioned generation.

Additionally, to highlight the limitations of the evaluation metrics, specifically, the optical flow-based endpoint error for the forward direction, EPE-F may be low if the action has not been performed between the start and the end image. In \Cref{fig:supp:intrep}, BAGEL-Do has a higher EPE compared to BAGEL-Do(SP), even though the action has been performed correctly in BAGEL-Do.

\myparagraph{Computational requirements.} We fine-tuned BAGEL on 4 A100 NVIDIA GPUs for $\approx 5$ hours. The zero-shot evaluations are performed on the same GPU set-up taking up to 2 hours to generate all metrics.
\section{User Study}
\label{sec:Userstudy}
We perform human evaluation to validate the performance of different models for action understanding. We provide the following instructions along with the input image; forward-generated image and the forward action; the reverse-generated image and the reverse action. We anonymize the model name and characteristics. We collected 240 diverse responses across 10 independent human evaluators. 
\begin{tcolorbox}[
    colback=blue!5!white, 
    colframe=blue!75!black, 
    title=\textbf{Instructions for User Study},
    fonttitle=\bfseries,
    coltitle=black, 
    colbacktitle=blue!20!white, 
    sharp corners,
    boxrule=0.5pt
]
    Consider the input image below and the action "turn on the tap" and the reverse action "turn off the tap". Based on the input image and the action/reverse action prompts. Rate the middle and the right image:
    \begin{itemize}
        \item Do not penalize the camera movements.
        \item Do not penalize image quality wrt to blur.
        \item Give full score if the action is completed or close to completion.
        \item When scoring, also consider the faithfulness of the object to the input image.
    \end{itemize}
\end{tcolorbox}

\begin{figure}[t]
\centering
    \includegraphics[width=0.8\linewidth]{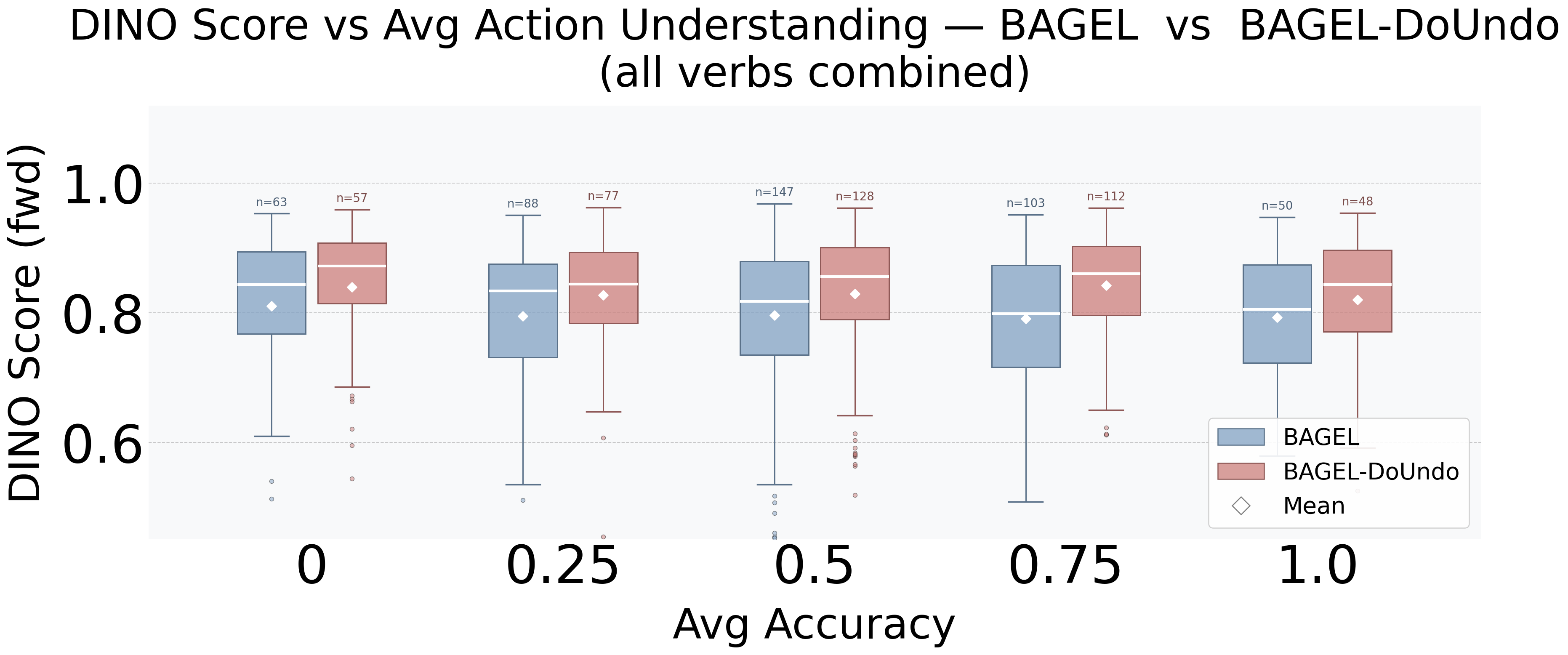}
    \captionof{figure}{\textbf{Quantitative results.} Distribution of action understanding and semantic awareness scores on the test set.}
    \label{fig:statescore}
    \vspace{-10pt}
  \end{figure}
\begin{table*}[t]
\centering
\smallskip
\scriptsize
\begin{tabularx}{\textwidth}{@{}XXXX@{}}
\toprule
Start Image & End Image  & BAGEL-Do  & BAGEL-Do(SP)\\
{\includegraphics[width=0.9\linewidth]{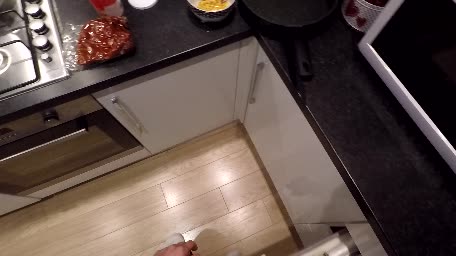}} & 
{\includegraphics[width=0.9\linewidth]{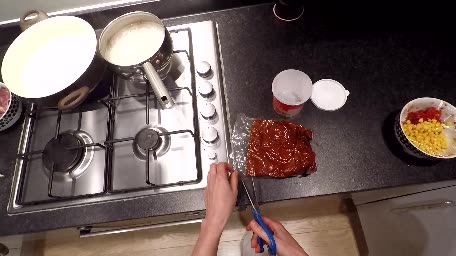}}
 &{\includegraphics[width=0.9\linewidth]{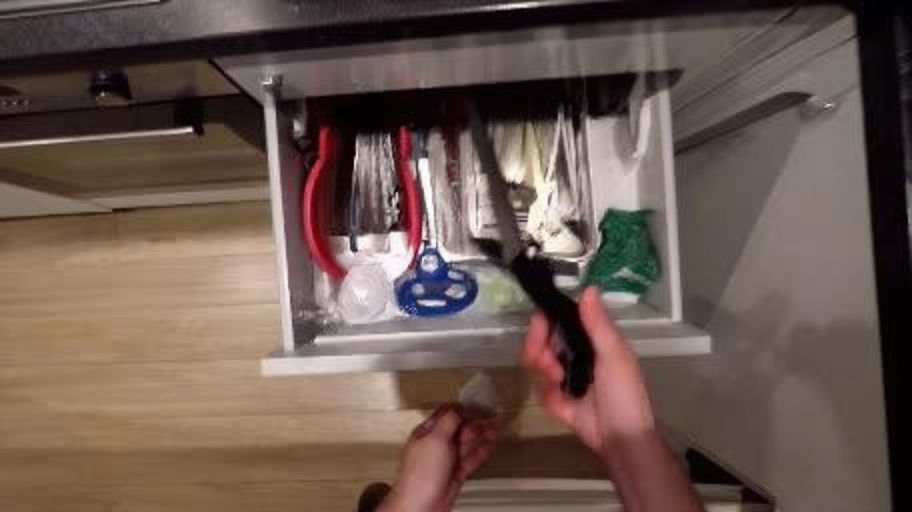}}
 &{\includegraphics[width=0.9\linewidth]{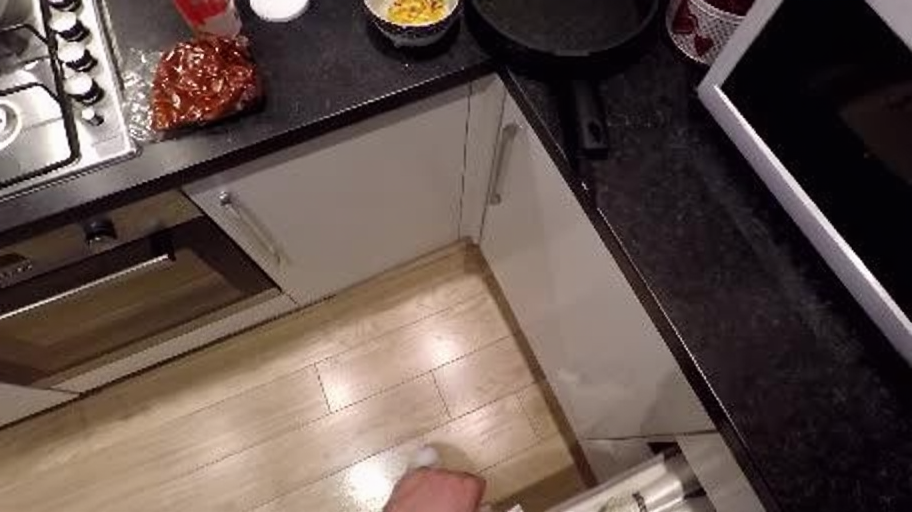}}
\\
& &   EPE: 238.88 & EPE: 143.96\\
 \multicolumn{4}{p{\textwidth}}{\textbf{Forward prompt:} The user is performing the action: 'get scissors'. The object is a pair of scissors, which is likely stored inside the lower kitchen cabinet directly below the countertop, as the cabinet door is partially open and the user's hand is reaching into it. The scissors are currently in a stored state, possibly on a shelf or drawer within the cabinet. The goal is to retrieve the scissors and bring them out of the cabinet to the countertop area for use. The user should use their right hand to grasp the scissors by the handles, ensuring a firm grip to prevent slipping.} \\
\bottomrule      
 \end{tabularx}
 \captionof{figure}{\textbf{Interpretation of the EPE metric.} A low EPE-F metric does not necessarily mean that the action has been performed. For the images generated using BAGEL-Do and BAGEL-Do(s); BAGEL-Do(SP), BAGEL-Do(SP) (col.~4) has lower EPE despite not performing the action.} 
\label{fig:supp:intrep}
\end{table*}

\section{Out-of-Domain Evaluation}
\label{sec:supp:ood}
\begin{figure*}[t]
    \centering
\includegraphics[width=\linewidth]{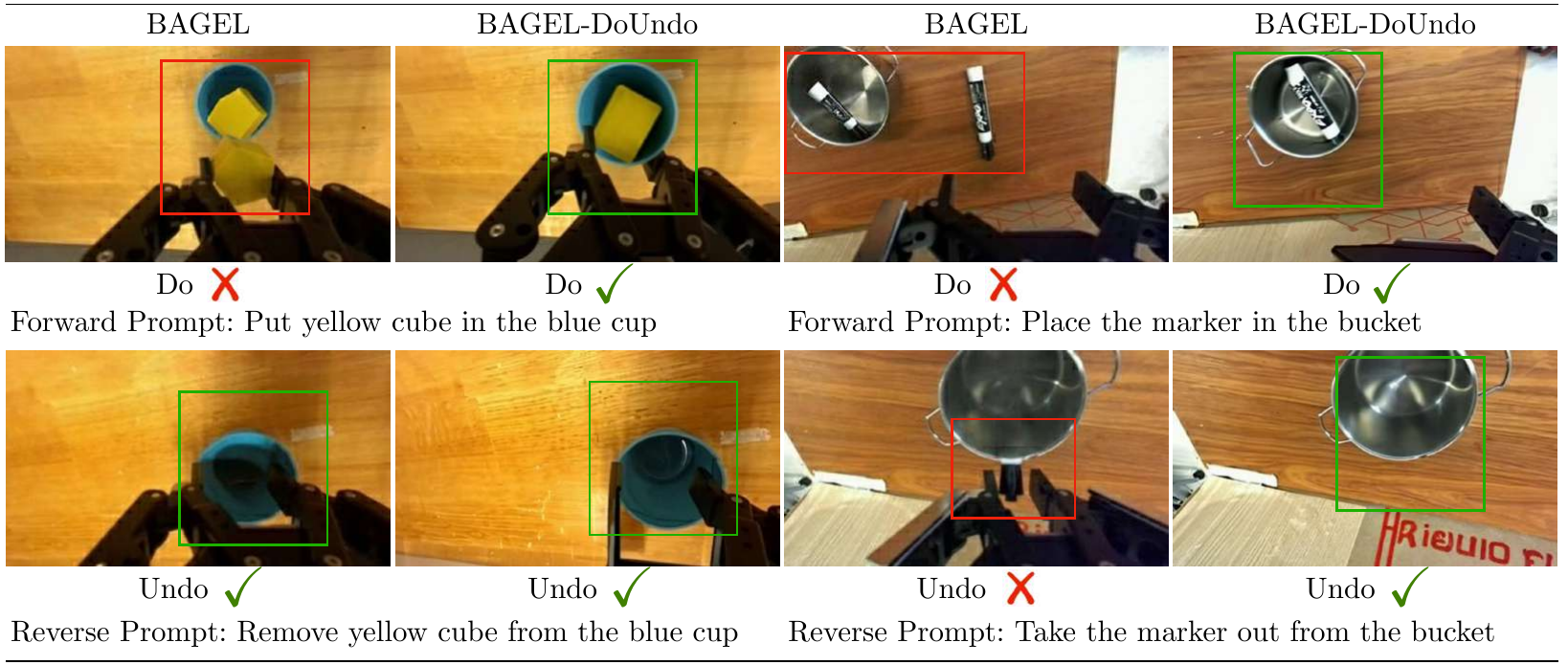}
    \caption{\textbf{Generalization to out-of-domain objects and environment.} BAGEL creates copies of the target yellow object and fails to remove the marker with unrealistic image generation on the Droid dataset \cite{DBLP:conf/rss/KhazatskyP0BDKN24}.}
    \label{fig:ood}
\end{figure*}

\begin{table*}[h!]
\centering
\smallskip
\scriptsize
\begin{tabularx}{\textwidth}{@{}XXXXX@{}}
\toprule
Input Image & \multicolumn{2}{c}{BAGEL} & \multicolumn{2}{c}{BAGEL-DoUndo}\\
\cmidrule(lr){2-3} \cmidrule(lr){4-5}
& Forward & Reverse & Forward & Reverse\\
{\includegraphics[width=0.9\linewidth]{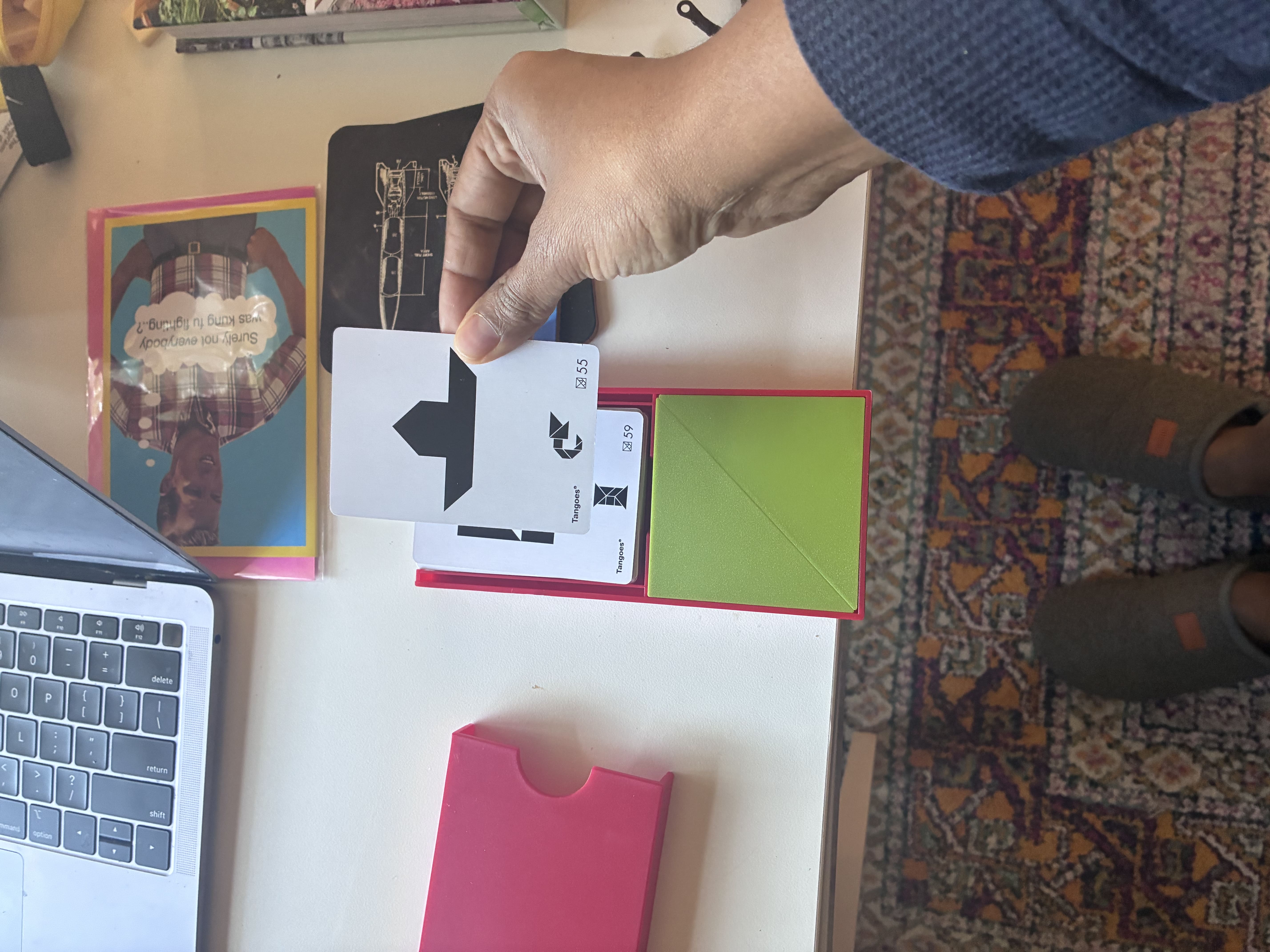}} & 
{\includegraphics[width=0.9\linewidth]{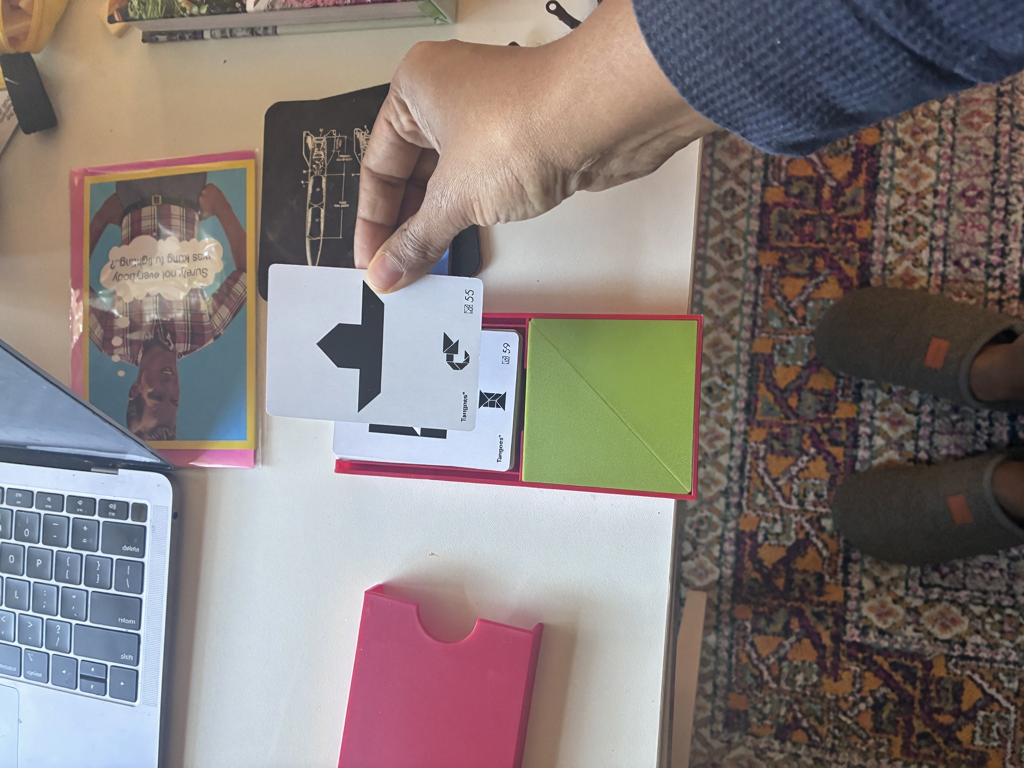}}
 &{\includegraphics[width=0.9\linewidth]{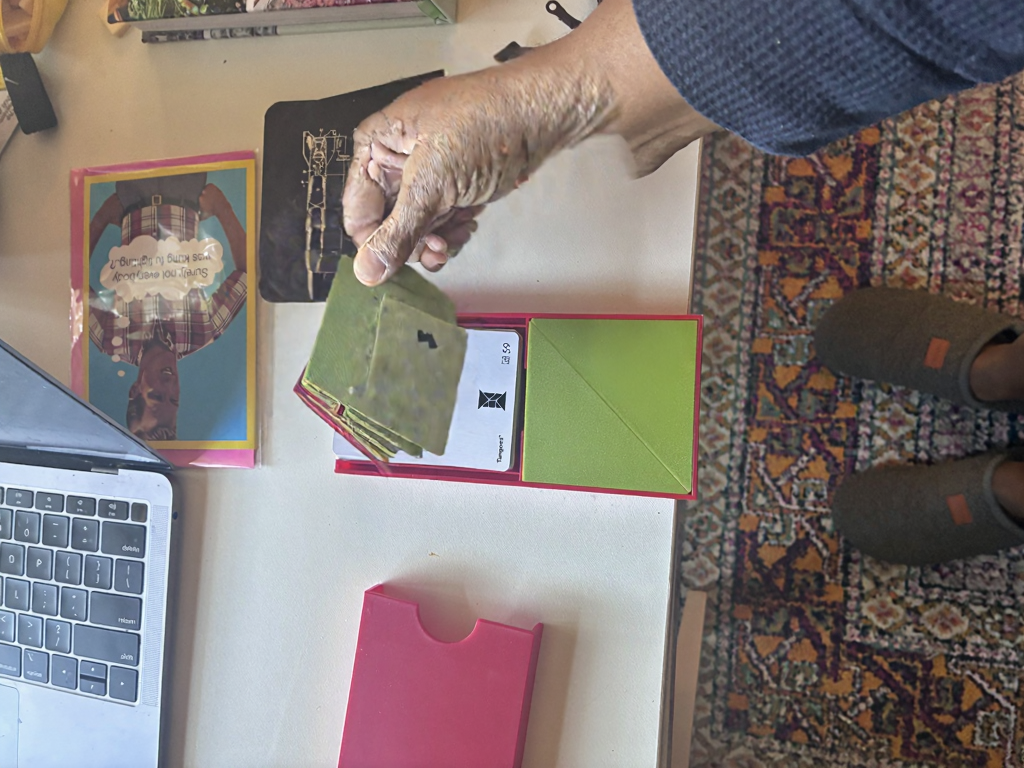}}
 &{\includegraphics[width=0.9\linewidth]{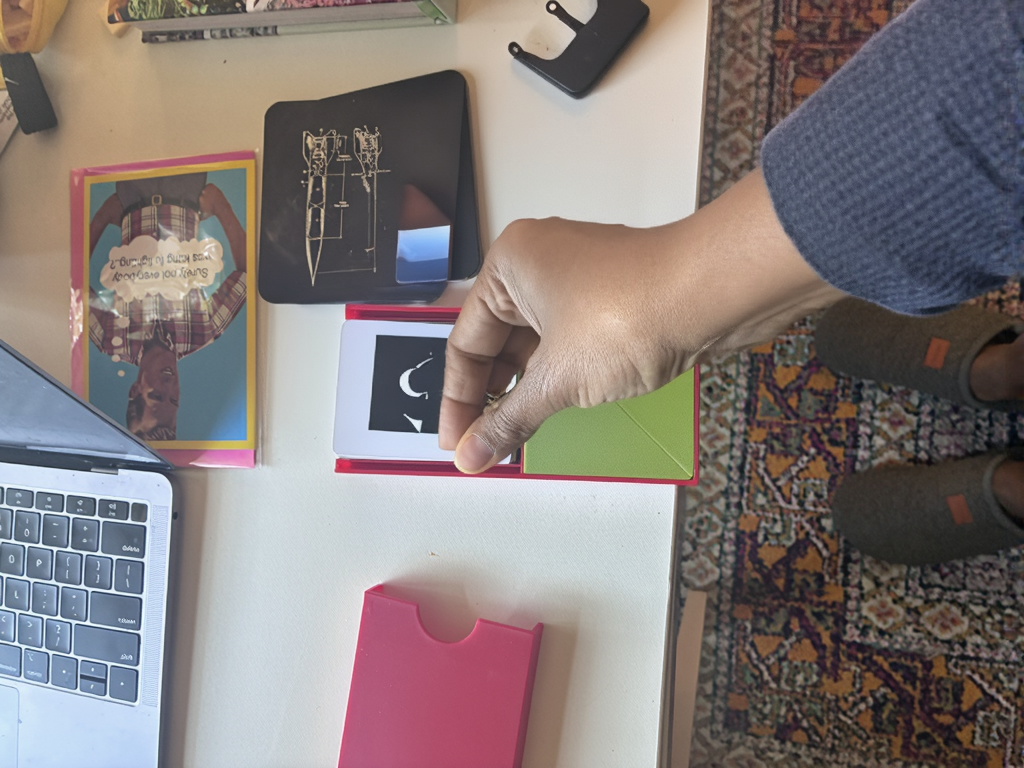}}
 & {\includegraphics[width=0.9\linewidth]{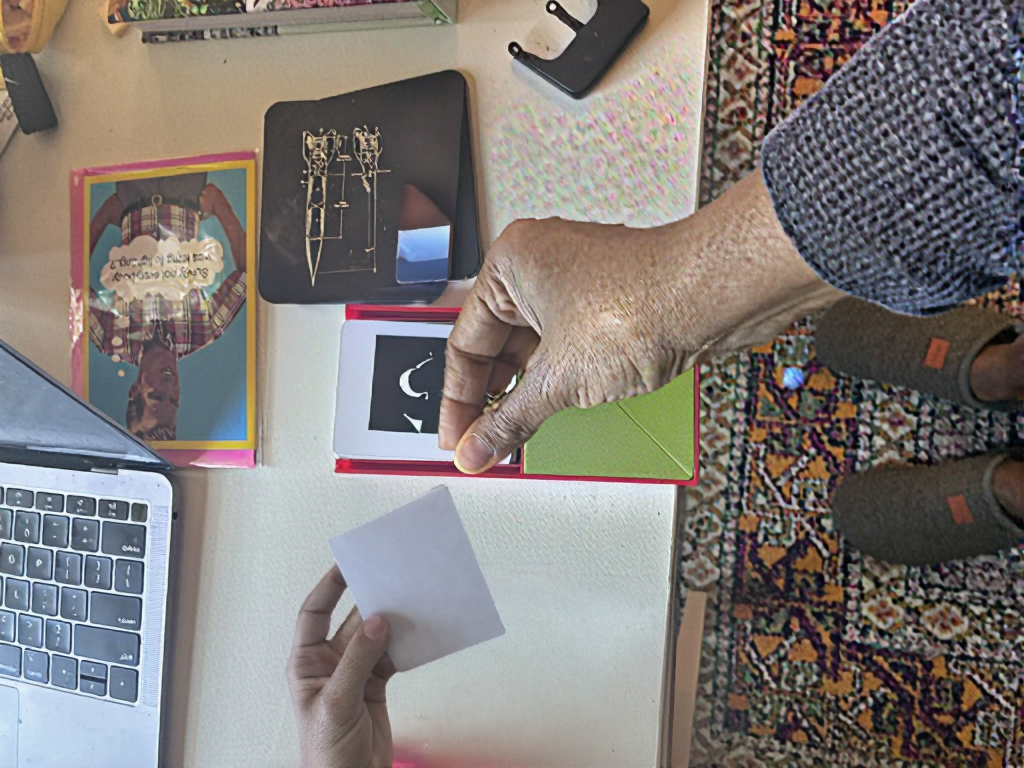}}
\\
 \multicolumn{5}{p{\textwidth}}{\textbf{Forward prompt:} Place the card held in the right hand onto the deck on the table, aligning it precisely with the existing stack of cards. 
 \newline
\textbf{Reverse prompt:} Grasp the top white and back card from the deck using a right-hand pinch grip and perform a vertical withdrawal to lift it clear of the stack. }\\
\midrule
{\includegraphics[width=0.9\linewidth]{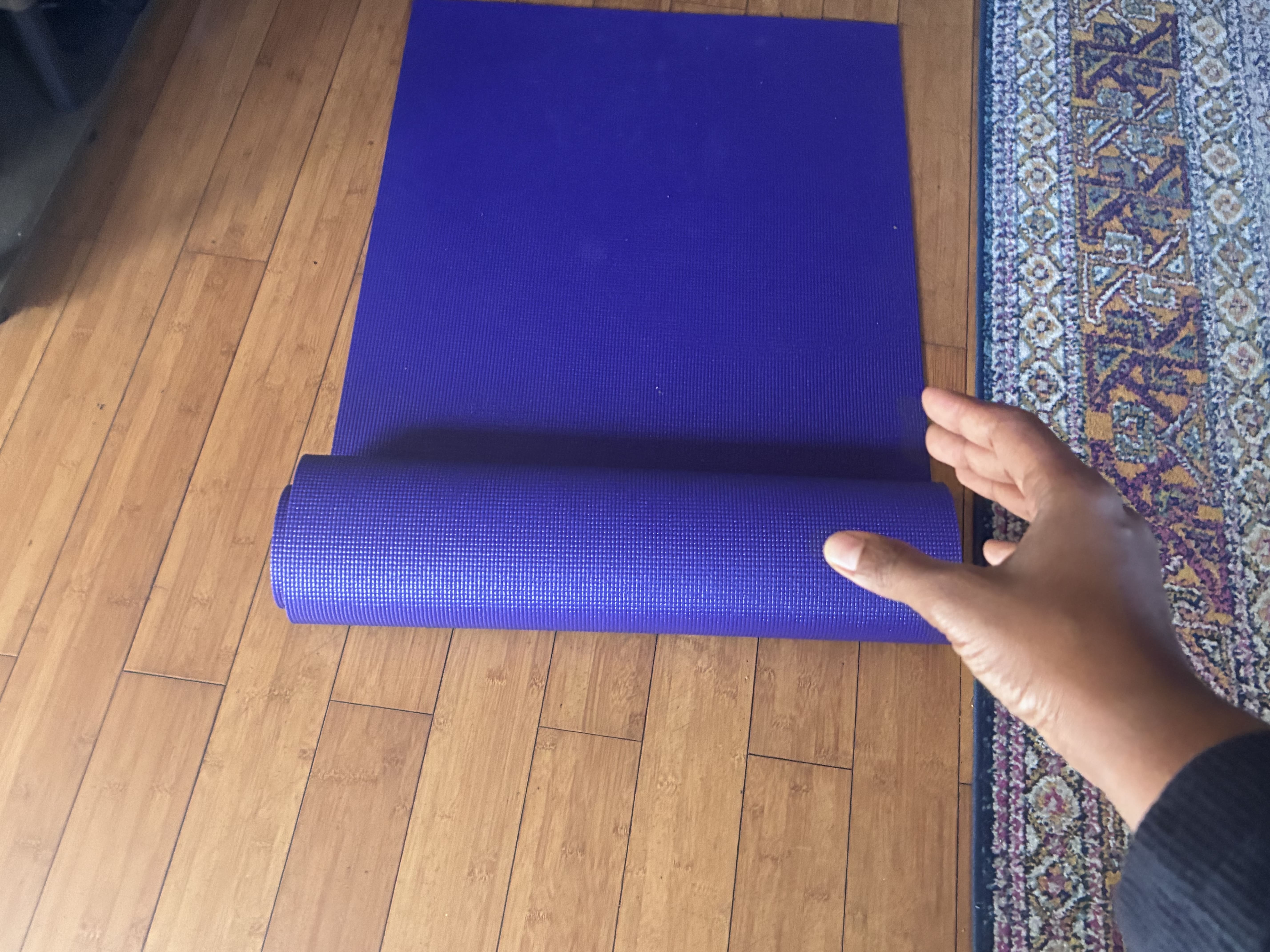}} & 
{\includegraphics[width=0.9\linewidth]{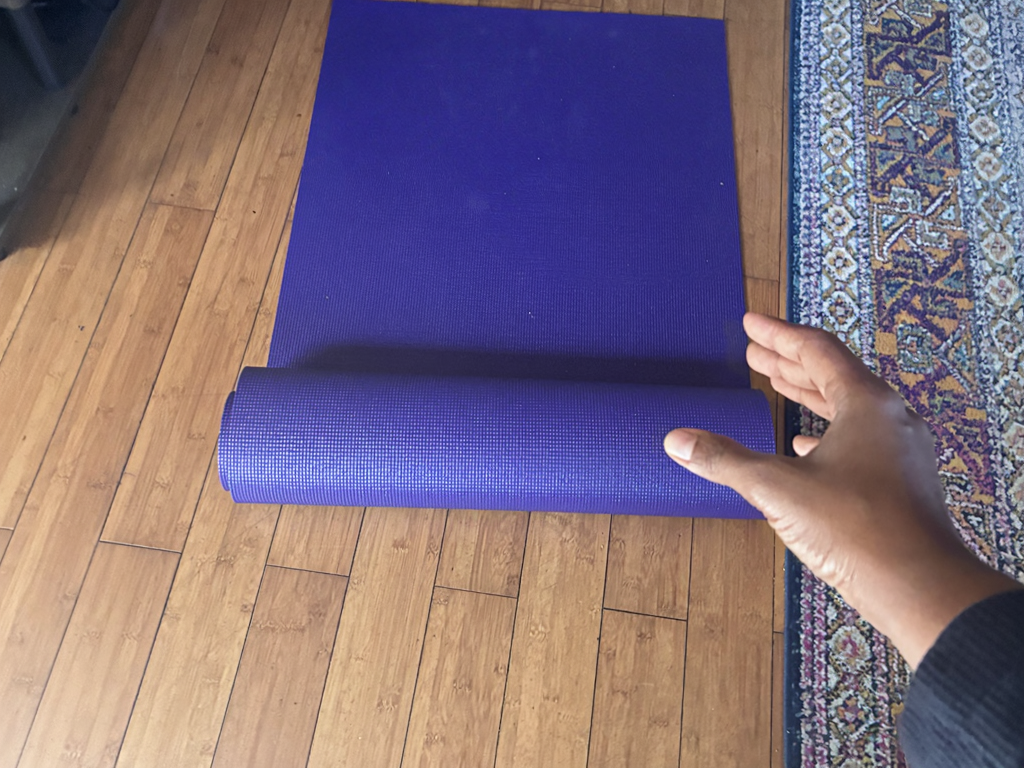}}
 &{\includegraphics[width=0.9\linewidth]{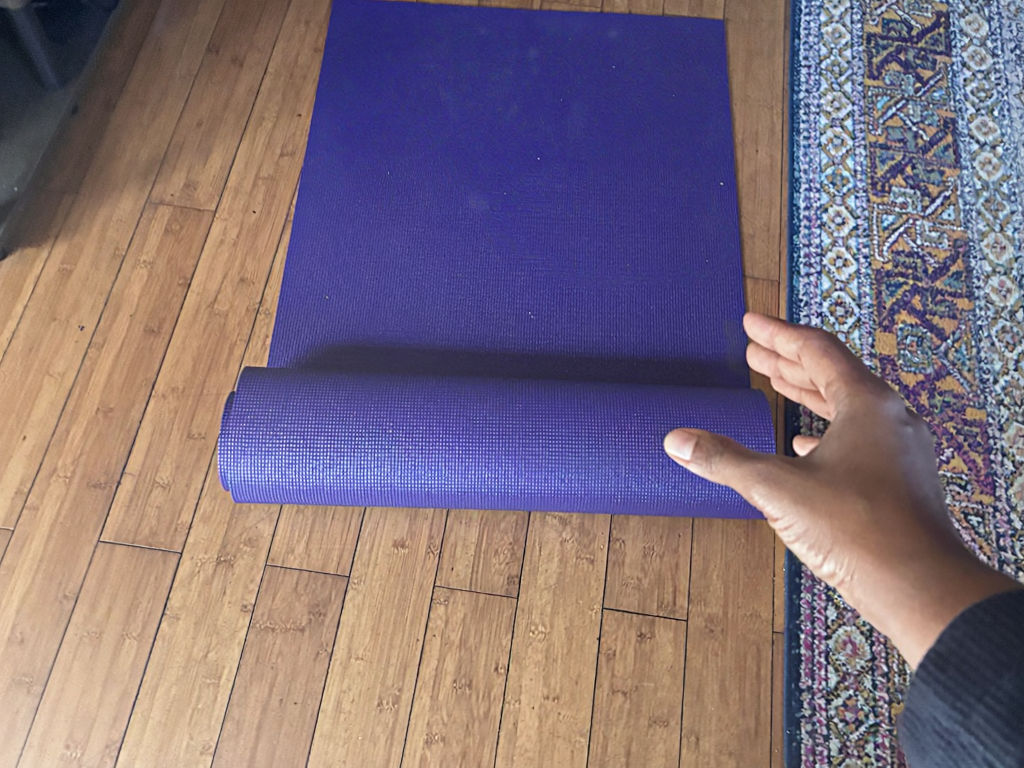}}
 &{\includegraphics[width=0.9\linewidth]{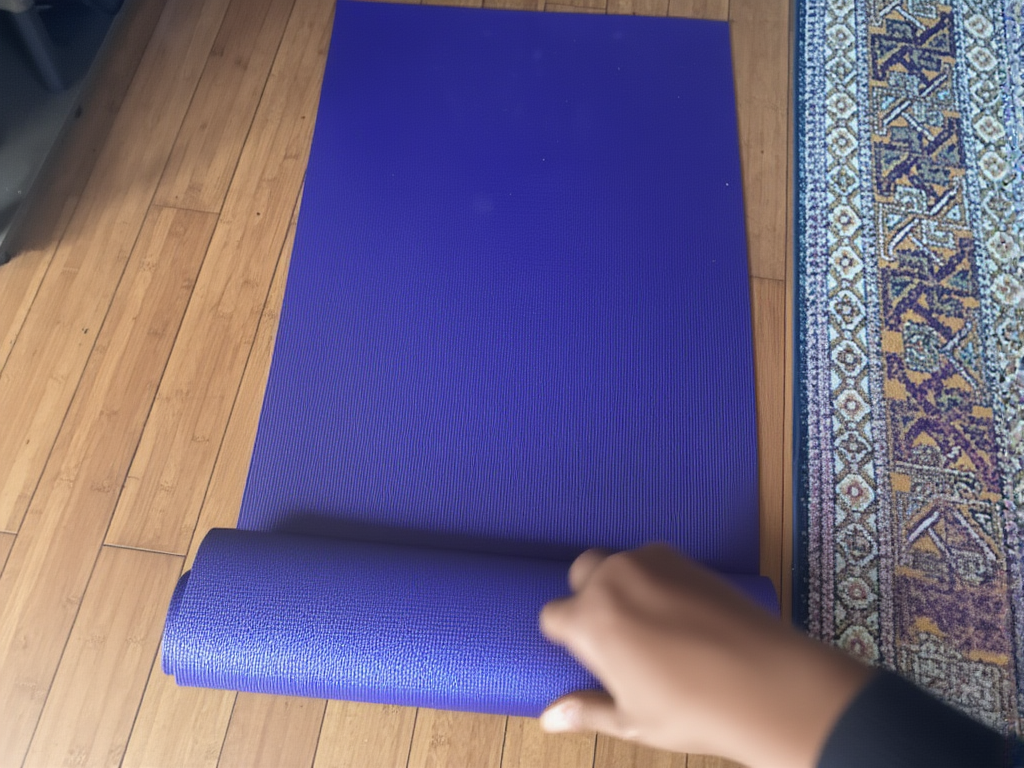}}
 & {\includegraphics[width=0.9\linewidth]{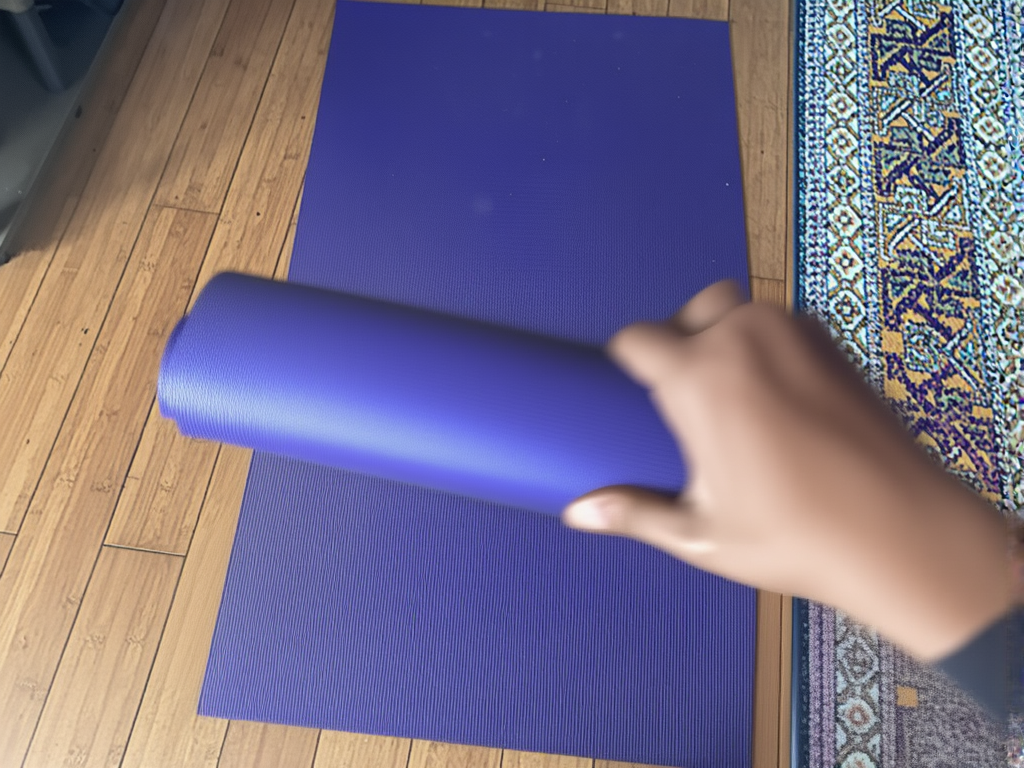}}
\\
\multicolumn{5}{p{\textwidth}}{\textbf{Forward prompt:} Use the right hand to push the rolled yoga mat outward away from the body until it lies completely flat on the floor.
\newline
\textbf{Reverse prompt:} Using a right-hand palmar grip, rotate the edge of the mat toward the body to form a tight, uniform cylinder revealing the wooden floor beneath.}\\
\midrule
{\includegraphics[width=0.9\linewidth]{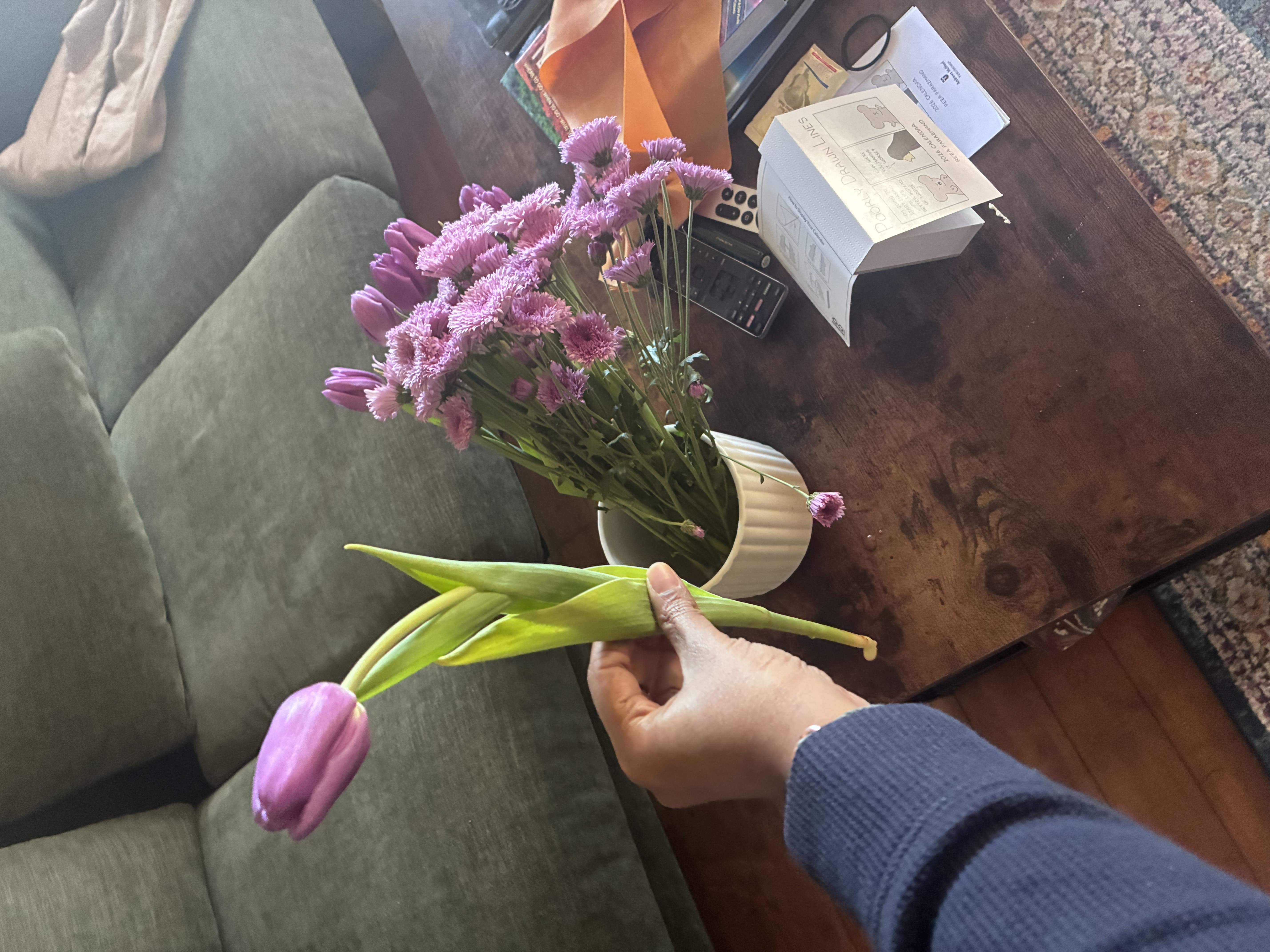}} & 
{\includegraphics[width=0.9\linewidth]{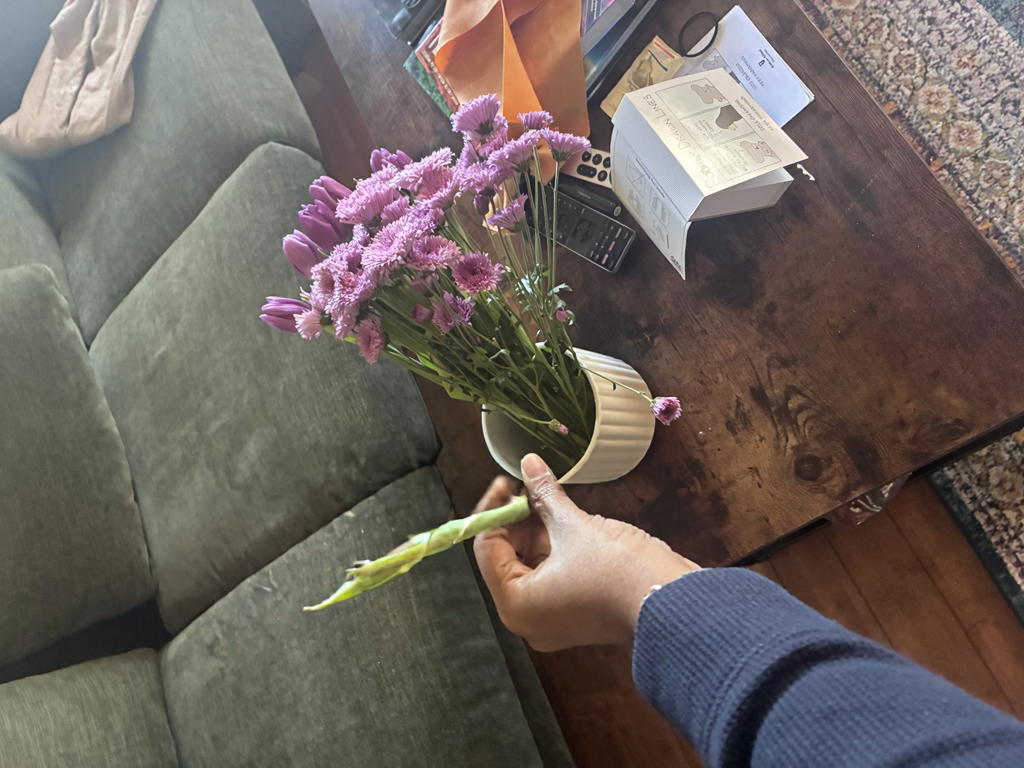}}
 &{\includegraphics[width=0.9\linewidth]{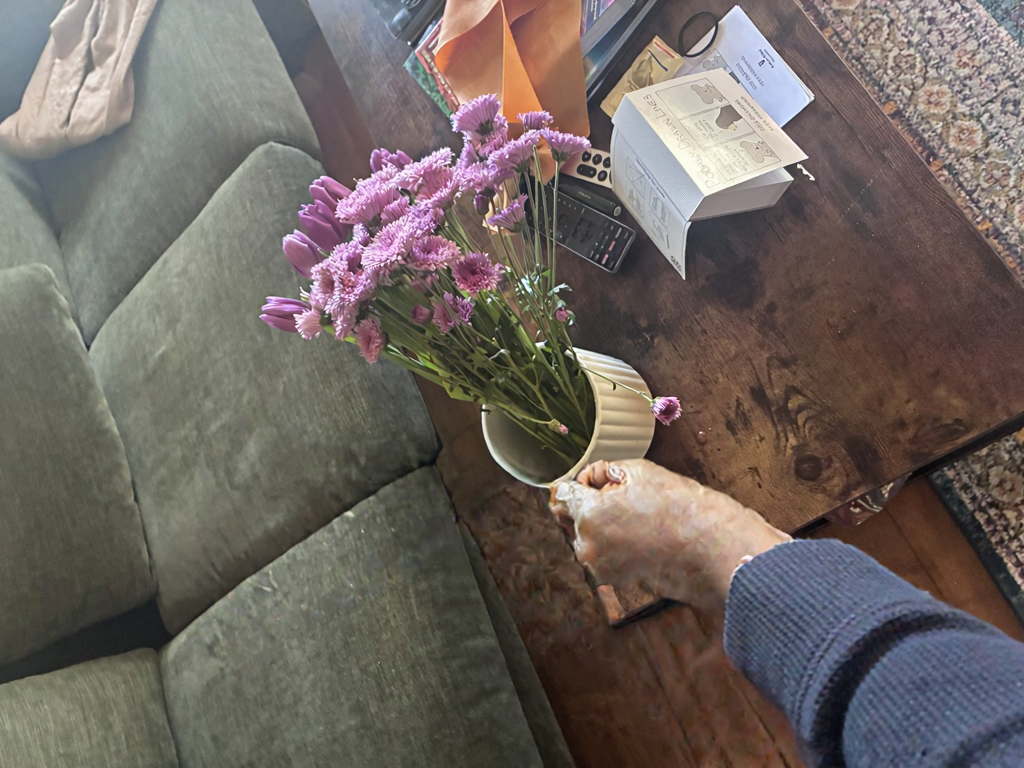}}
 &{\includegraphics[width=0.9\linewidth]{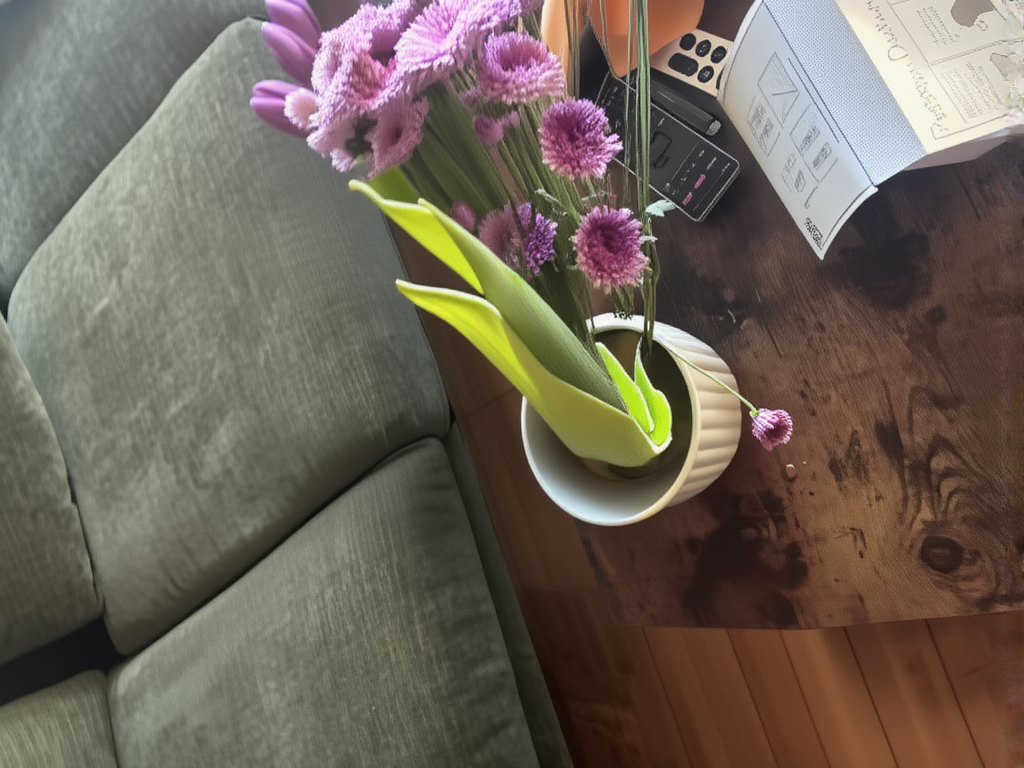}}
 & {\includegraphics[width=0.9\linewidth]{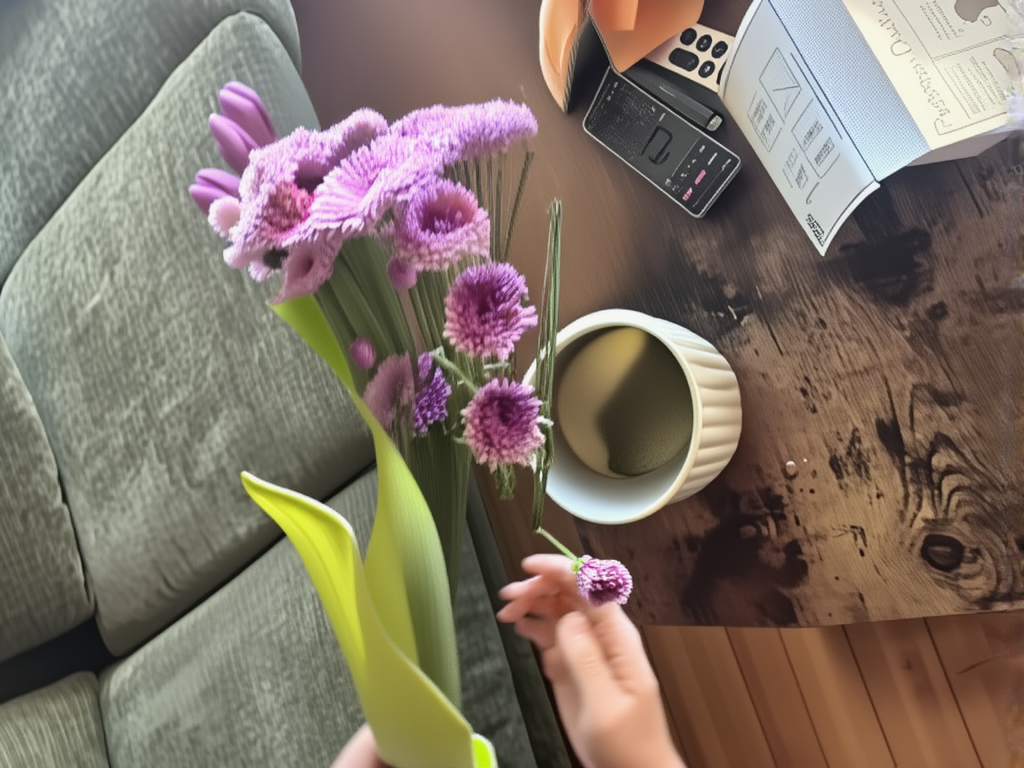}}
\\
\multicolumn{5}{p{\textwidth}}{\textbf{Forward prompt:} Grasp the purple tulip stem and insert it into the center of the white fluted vase, making contact with the bottom surface, and aligning it with the existing cluster of purple chrysanthemums.
\newline
\textbf{Reverse prompt:} Grasp a single purple tulip stem from the cluster and perform a vertical withdraw, pulling it upward until it is completely clear of the vase rim.
}\\
\midrule
{\includegraphics[width=0.9\linewidth]{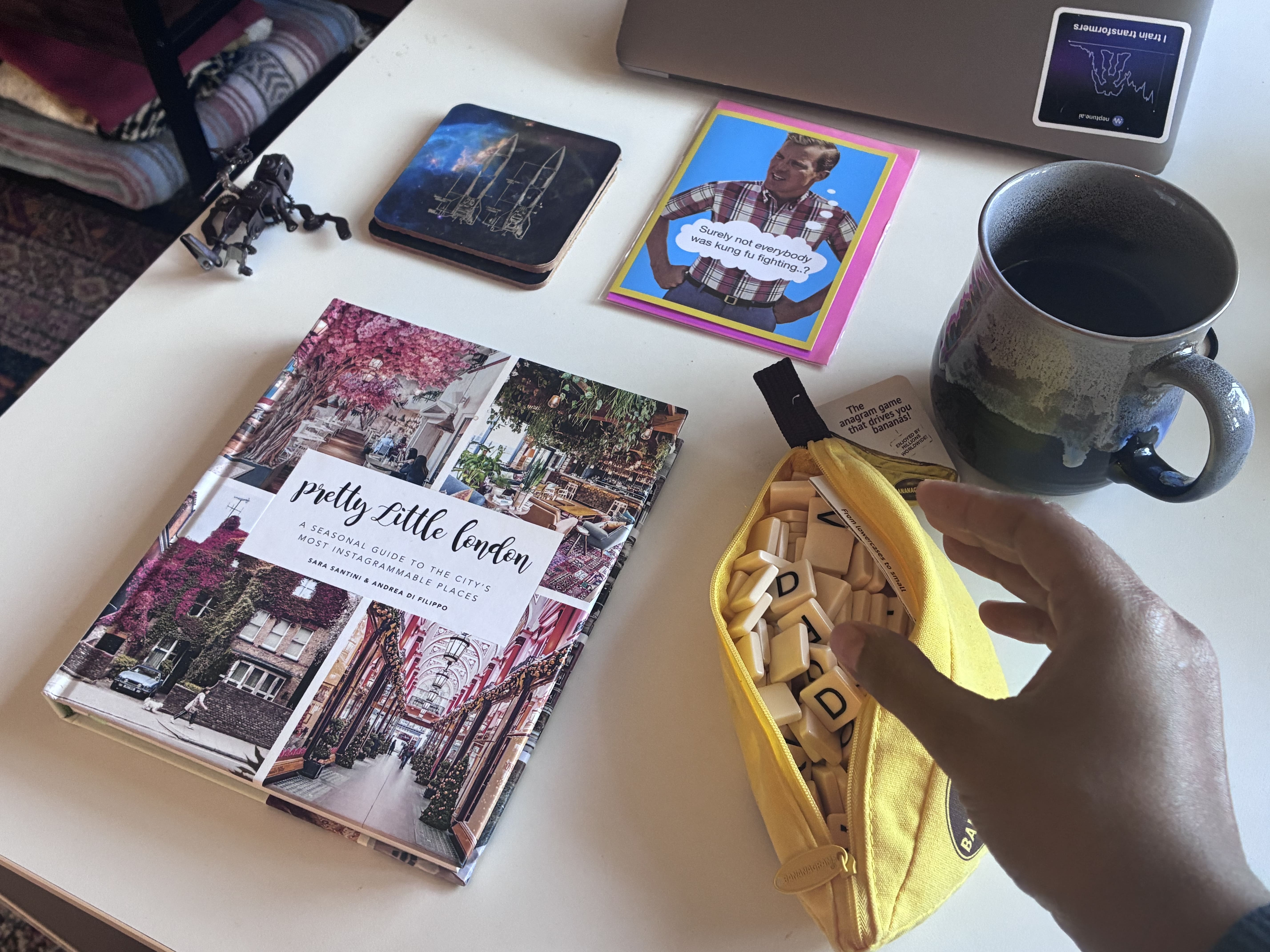}} & 
{\includegraphics[width=0.9\linewidth]{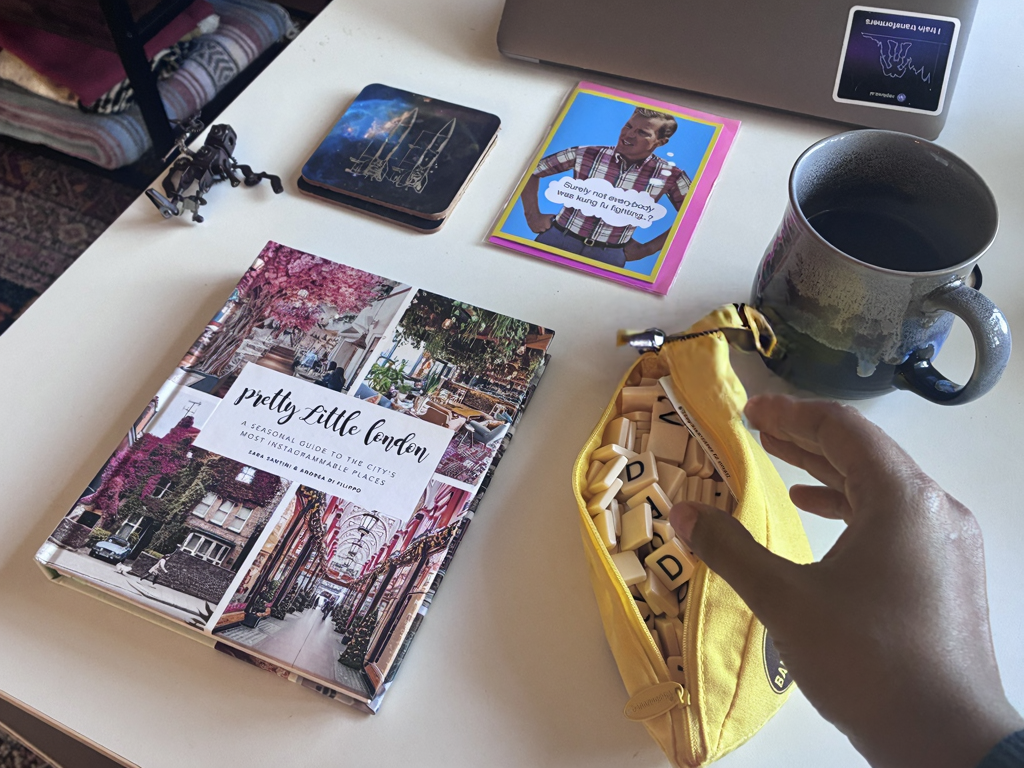}}
 &{\includegraphics[width=0.9\linewidth]{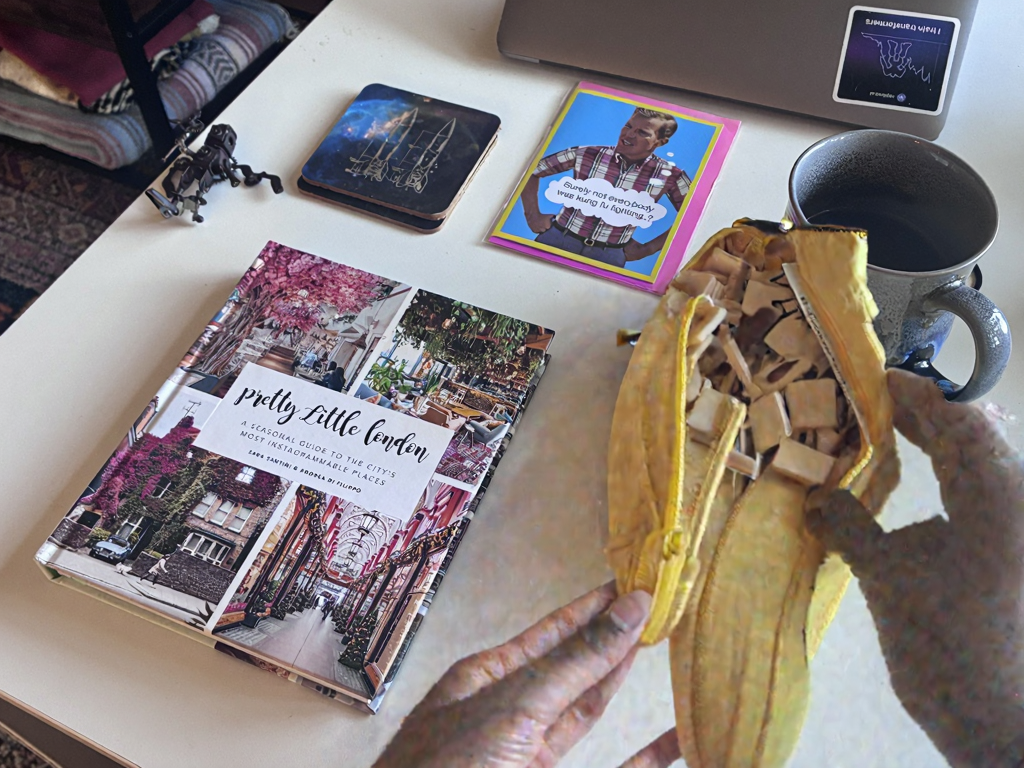}}
 &{\includegraphics[width=0.9\linewidth]{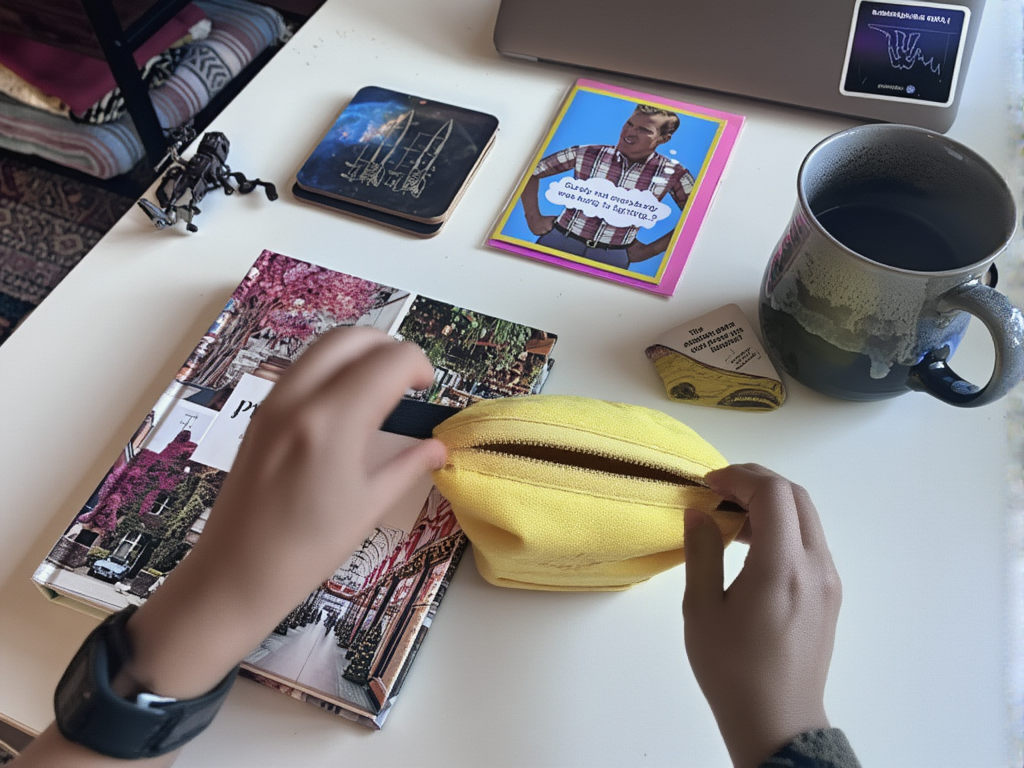}}
 & {\includegraphics[width=0.9\linewidth]{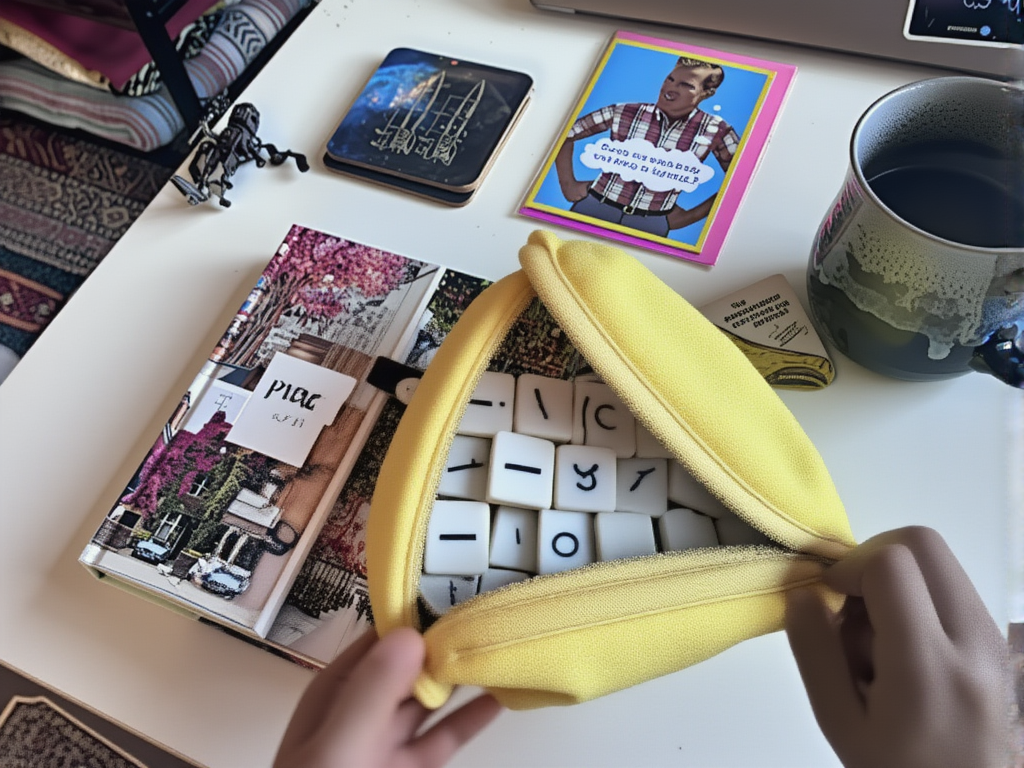}}
\\
\multicolumn{5}{p{\textwidth}}{\textbf{Forward prompt:} close the zip of the pouch by pinching the yellow zipper tab and pulling it away from the body to fully seal the yellow pouch on the table.
\newline
\textbf{Reverse prompt:} open the banana-shaped pouch by pulling the yellow zipper tab toward the body to reveal the small white plastic letter tiles inside.
}\\
\bottomrule      
 \end{tabularx}
 \captionof{figure}{\textbf{Out of domain and in the wild evaluation.} Qualitative comparison BAGEL \cite{deng2025bagel} with BAGEL-DoUndo on real-world actions and objects that are not present in the Do-Undo training data or benchmark.} 
\label{fig:supp:inthewild}
\end{table*}
To show the generalization abilities of our task and the trained BAGEL-DoUndo in the wild scenarios, we manually curate qualitative examples in \Cref{fig:supp:inthewild} and show comparative performance against BAGEL. 
We show diverse objects such as cards, a yoga mat, a flower, and a pouch, with diverse actions such as push, grasp, and zip, which are not present in our training data.
Here again, the model trained on the DoUndo dataset consistently outperforms the baseline, supporting our training hypothesis for action-grounded generation within the DoUndo paradigm.

\section{Additional Qualitative Examples}
\label{sec:supp:additionalqualitatives}
In \cref{fig:supp:multiturn}, we demonstrate an example of evolutionary action generation to analyze the context alignment and the action grounding abilities of a unified VLM. Starting from an image and an input prompt with action description, we generate an action conditioned image which is subsequently used as a state on which an action is performed. In multi-turn setting, all the previous and current prompt, including the previously generated images are provided as context.
Here again, we observe that BAGEL-DoUndo generates action and background consistent images during the four-step generation process.

In \Cref{fig:supp:qualitative}, we provide additional qualitative examples for action-conditioned generation.
The images include objects and actions that are not present in the training domain of the Do-Undo dataset.

\begin{table*}[t]
\smallskip
\scriptsize
\begin{tabularx}{\textwidth}{@{}m{0.25cm} *{4}{>{\centering\arraybackslash}X@{}}}
\toprule
{\rotatebox[origin=t]{90}{Start Image}}& {\includegraphics[width=0.9\linewidth,height=0.7\linewidth]{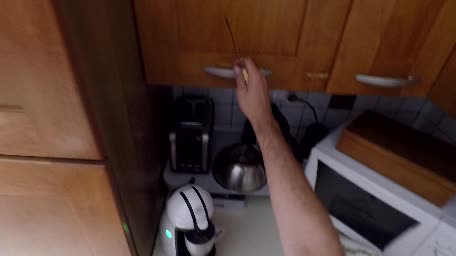}} & & &\\
& Step 1 & Step 2 & Step 3 & Step 4\\
\multirow{2}{*}{\rotatebox{90}{BAGEL}}&{\includegraphics[width=0.9\linewidth,height=0.7\linewidth]{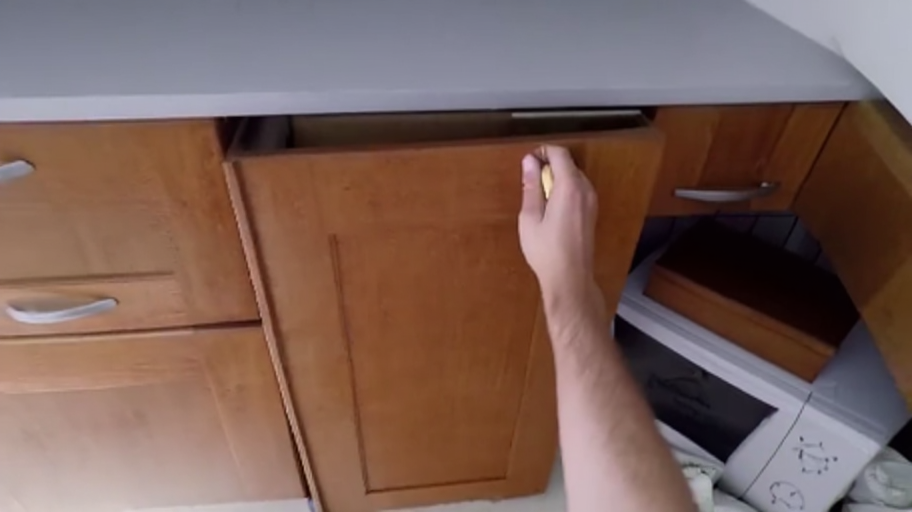}}
&{\includegraphics[width=0.9\linewidth,height=0.7\linewidth]{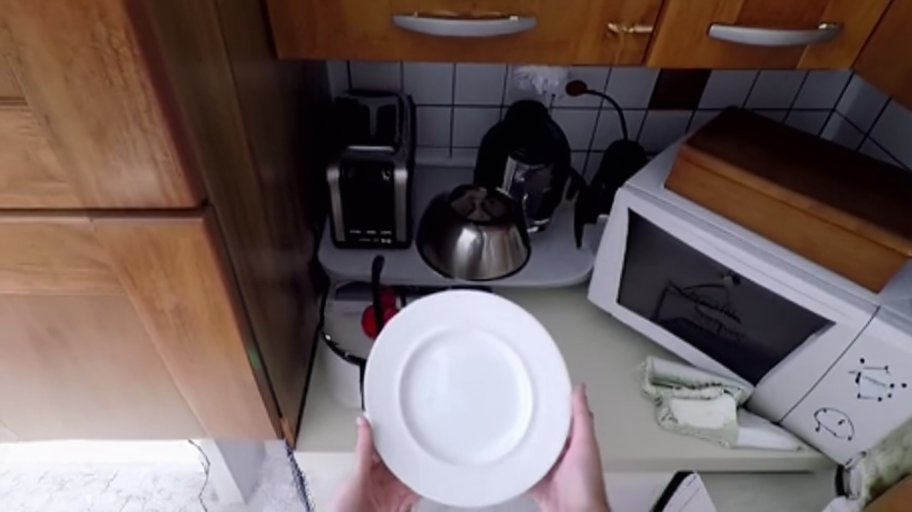}}
&{\includegraphics[width=0.9\linewidth,height=0.7\linewidth]{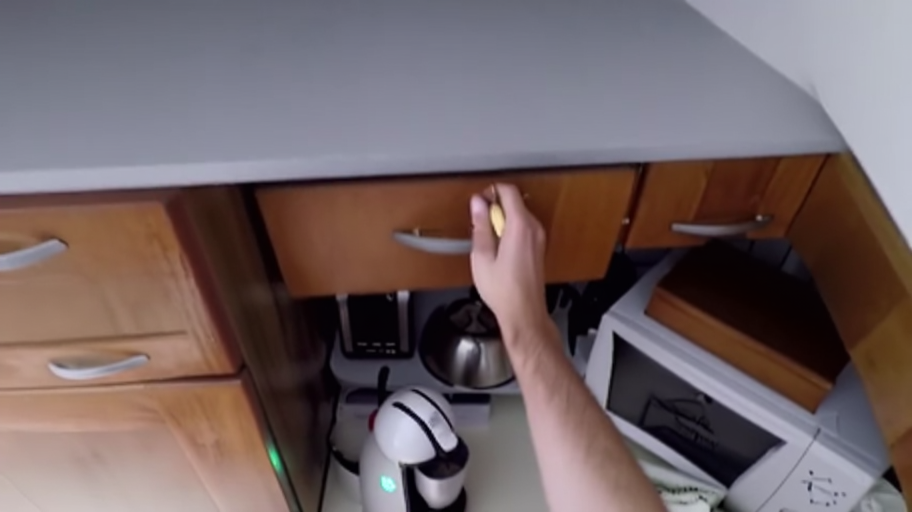}}
&
{\includegraphics[width=0.9\linewidth,height=0.7\linewidth]{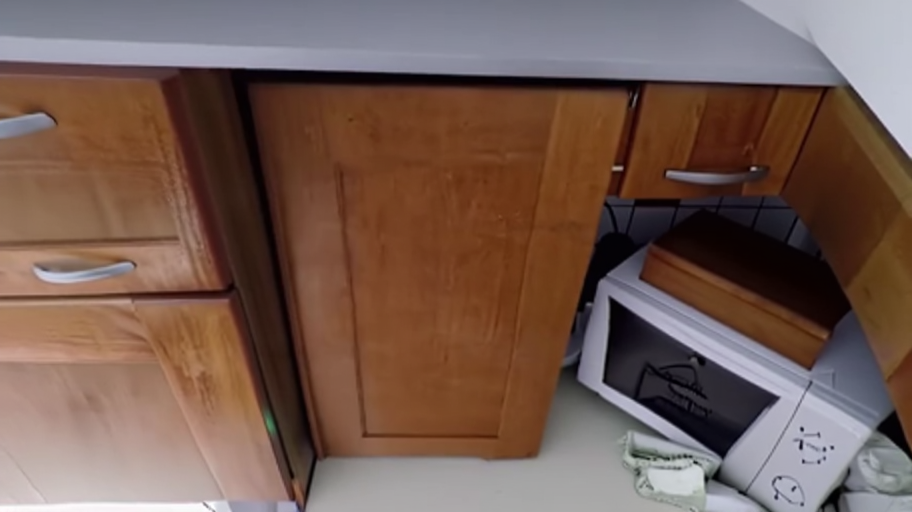}}\\
&{\includegraphics[width=0.9\linewidth,height=0.7\linewidth]{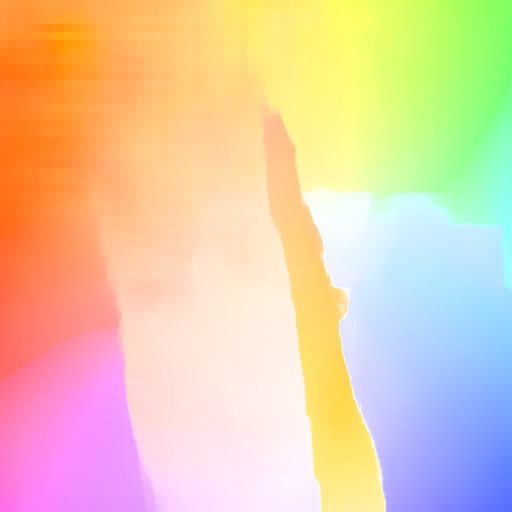} }
&{\includegraphics[width=0.9\linewidth,height=0.7\linewidth]{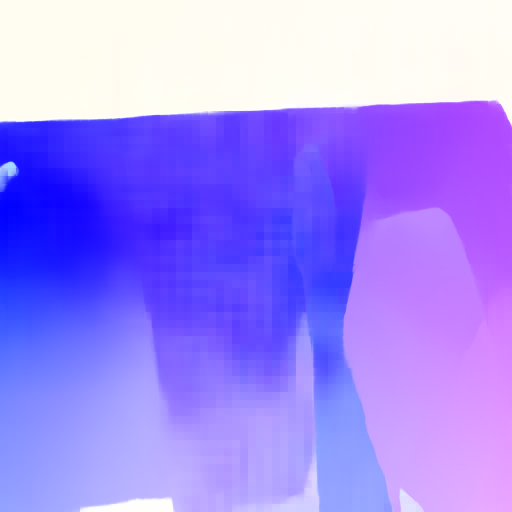} }
&{\includegraphics[width=0.9\linewidth,height=0.7\linewidth]{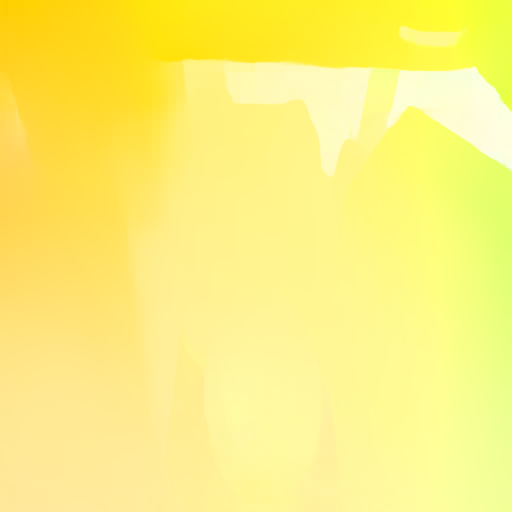} }
&{\includegraphics[width=0.9\linewidth,height=0.7\linewidth]{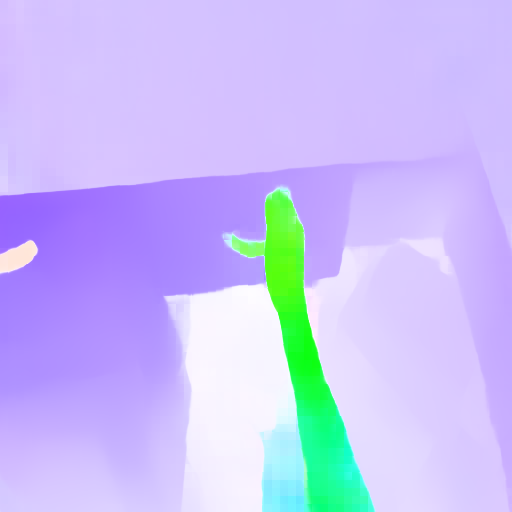} } \\
\multirow{2}{*}{\rotatebox{90}{BAGEL-DoUndo}}&{\includegraphics[width=0.9\linewidth,height=0.7\linewidth]{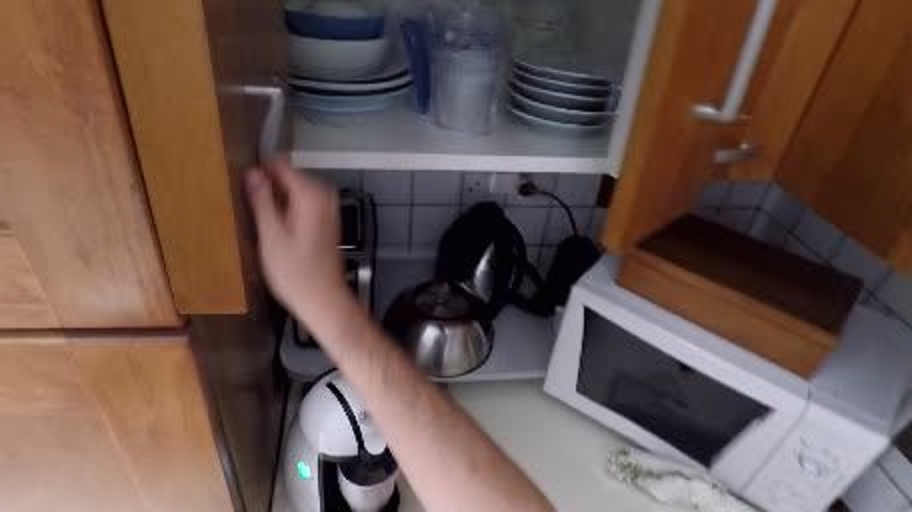} }
&{\includegraphics[width=0.9\linewidth,height=0.7\linewidth]{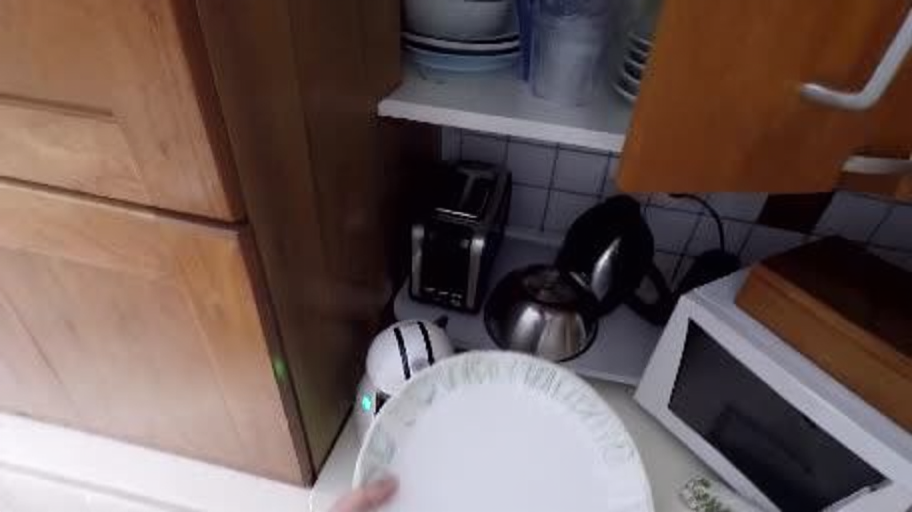} }
&{\includegraphics[width=0.9\linewidth,height=0.7\linewidth]{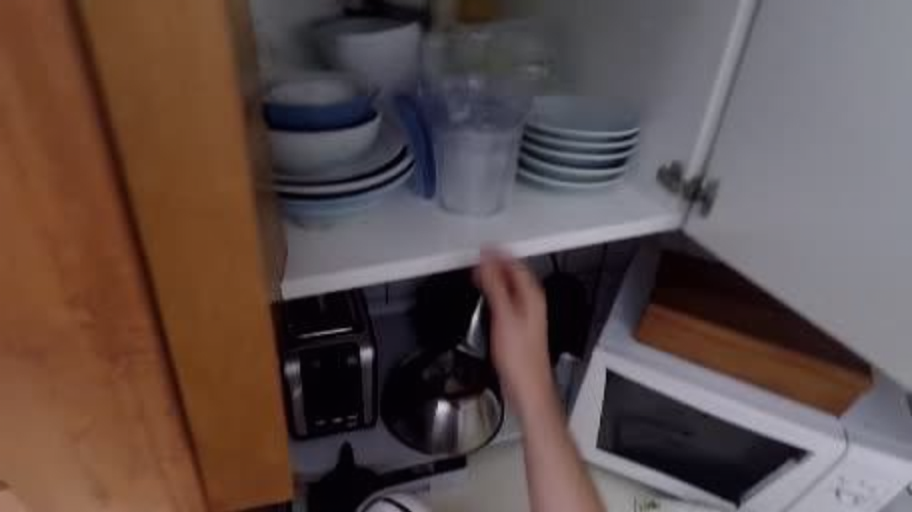} }
&{\includegraphics[width=0.9\linewidth,height=0.7\linewidth]{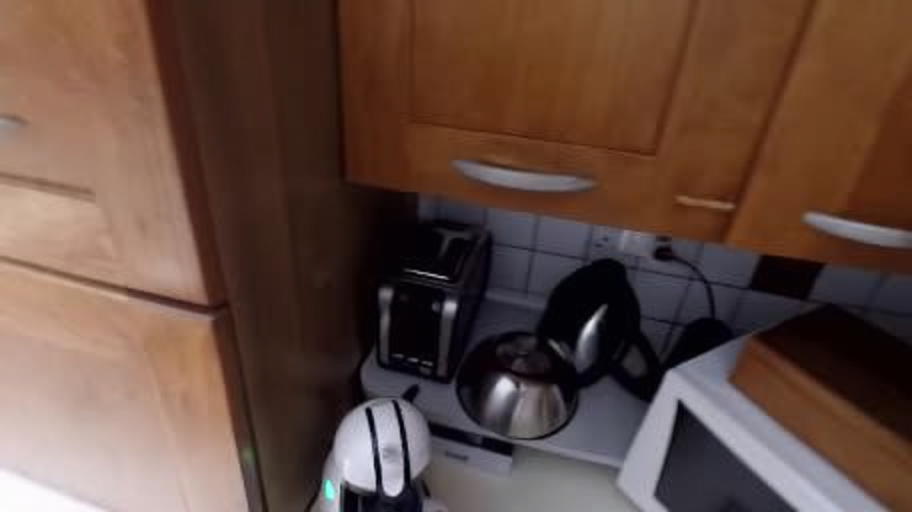} }\\
&{\includegraphics[width=0.9\linewidth,height=0.7\linewidth]{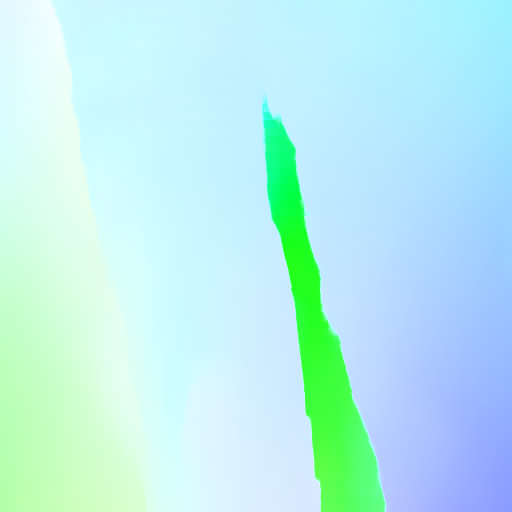} }
&{\includegraphics[width=0.9\linewidth,height=0.7\linewidth]{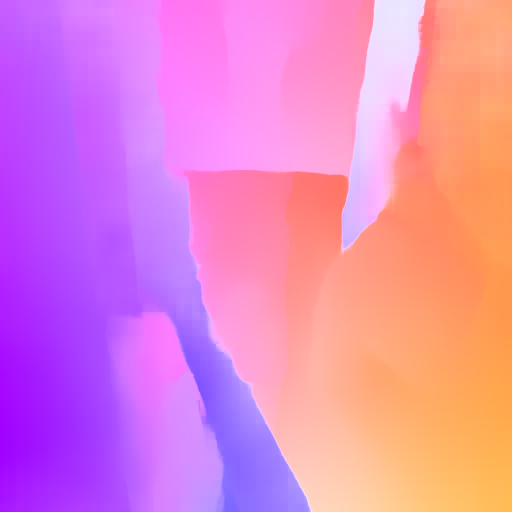} }
&{\includegraphics[width=0.9\linewidth,height=0.7\linewidth]{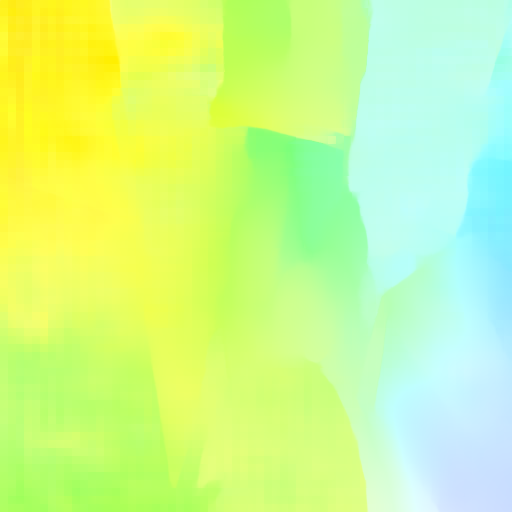} }
&{\includegraphics[width=0.9\linewidth,height=0.7\linewidth]{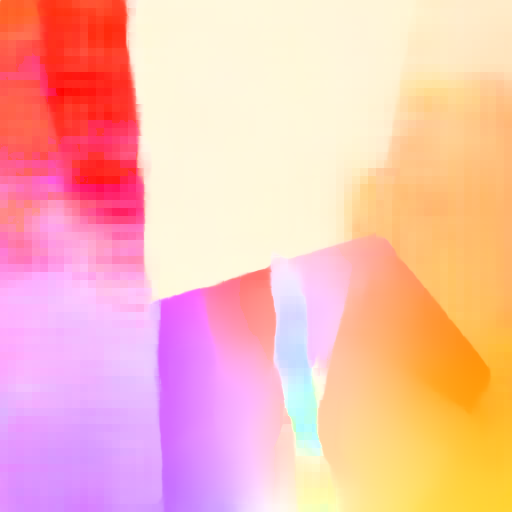} } \\
& \parbox[t]{0.9\linewidth}{
        {Open the wooden kitchen cabinet door located above the countertop, to the left of the microwave.Use your right hand to grasp the handle firmly and pull it outward to open the door}}
& \parbox[t]{0.9\linewidth}{
        {Pick up the top white ceramic plate from the stack on the lower shelf of the kitchen cabinet positioned above the kettle while maintaining the cabinet door in its current fully open position}}
& \parbox[t]{0.9\linewidth}{
        {Place the white ceramic plate back onto the stack on the lower shelf of the open kitchen cabinet. The plate is currently in a handheld state, and the objective is to return it to a stable, resting position at the top of the white plate stack inside the wooden cabinet. 
        Align the plate directly over the existing stack on the shelf and lower it steadily until it rests flat and secure.}}
&\parbox[t]{0.9\linewidth}{
        {Close the wooden kitchen cabinet door located above the countertop, to the left of the microwave.}}
\\
\bottomrule      
 \end{tabularx}
 \captionof{figure}{\textbf{Multi-turn generation with evolving actions.} We show the ability of BAGEL-DoUndo to perform actions in a multi-turn fashion. Starting from a start state image, the model performs a series of actions conditioned on the previous generated state.} 
\label{fig:supp:multiturn}
\end{table*}
\begin{table*}[t]
\centering
\smallskip
\scriptsize
\begin{tabularx}{\textwidth}{@{}XXXXX@{}}
\toprule
Input Image & \multicolumn{2}{c}{BAGEL} & \multicolumn{2}{c}{BAGEL-DoUndo}\\
\cmidrule(lr){2-3} \cmidrule(lr){4-5}
& Forward & Reverse & Forward & Reverse\\
{\includegraphics[width=\linewidth]{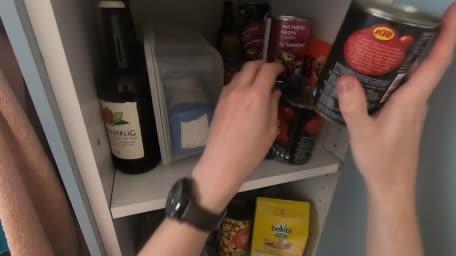}} & 
{\includegraphics[width=\linewidth]{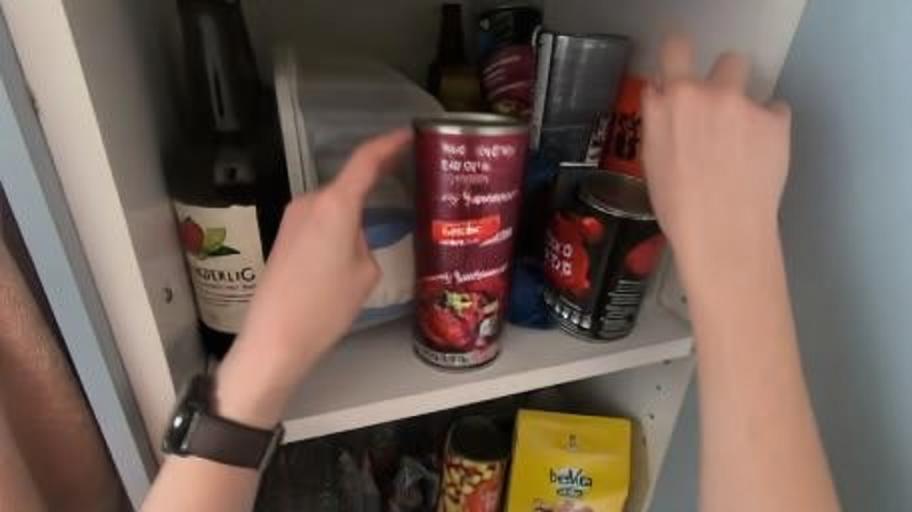}}
 &{\includegraphics[width=\linewidth]{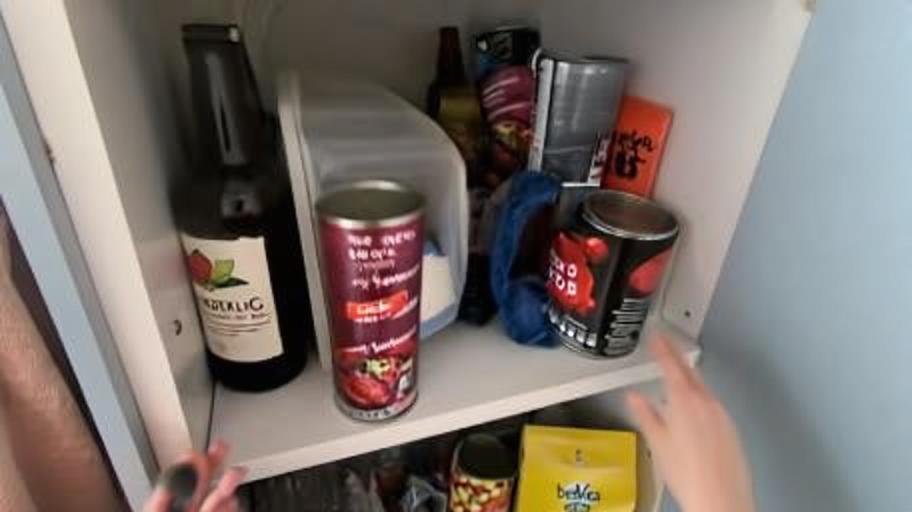}}
 &{\includegraphics[width=\linewidth]{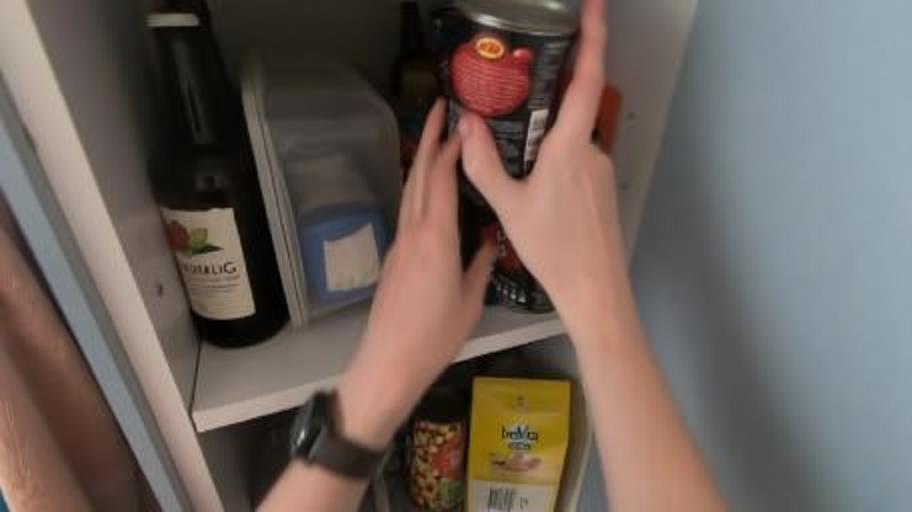}}
 & {\includegraphics[width=\linewidth]{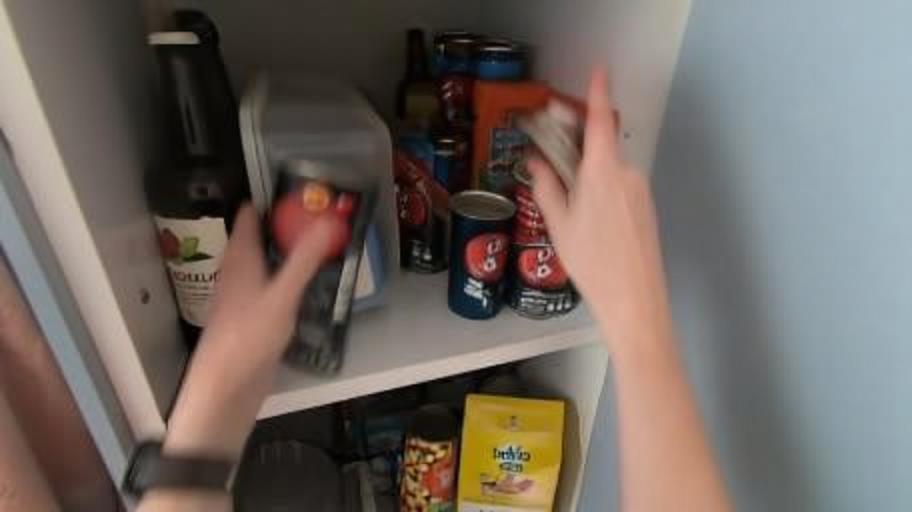}}
\\
\multicolumn{5}{p{\textwidth}}{\textbf{Forward prompt:} stack the red cylindrical tomato cans on the shelf with the right hand while holding another can with the left hand. \newline \textbf{Reverse prompt:} remove the red cylindrical tomato cans from the shelf with the right hand while holding another can with the left hand.}
\\
\midrule
{\includegraphics[width=\linewidth]{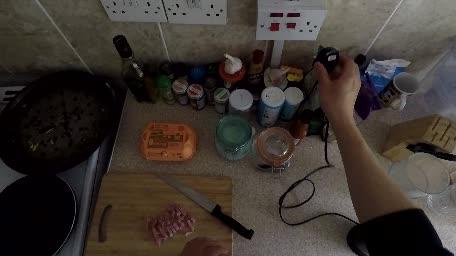}} & 
{\includegraphics[width=\linewidth]{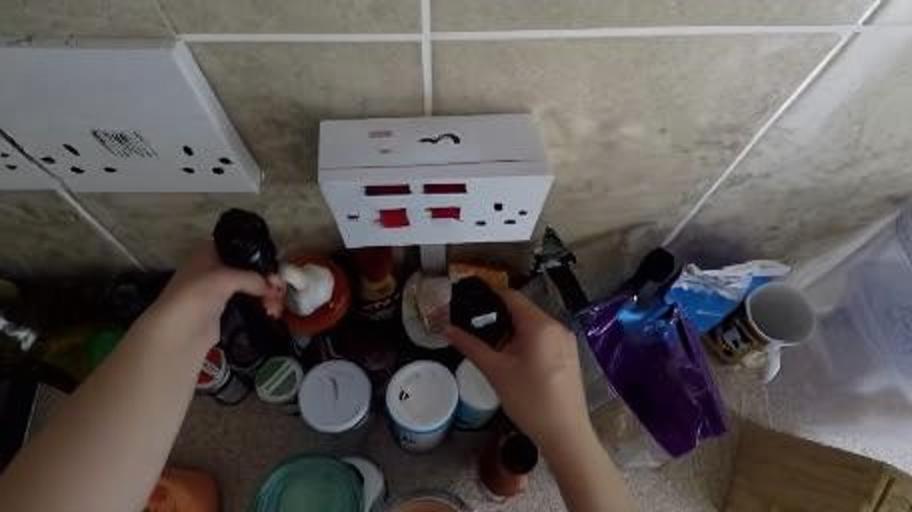}}
 &{\includegraphics[width=\linewidth]{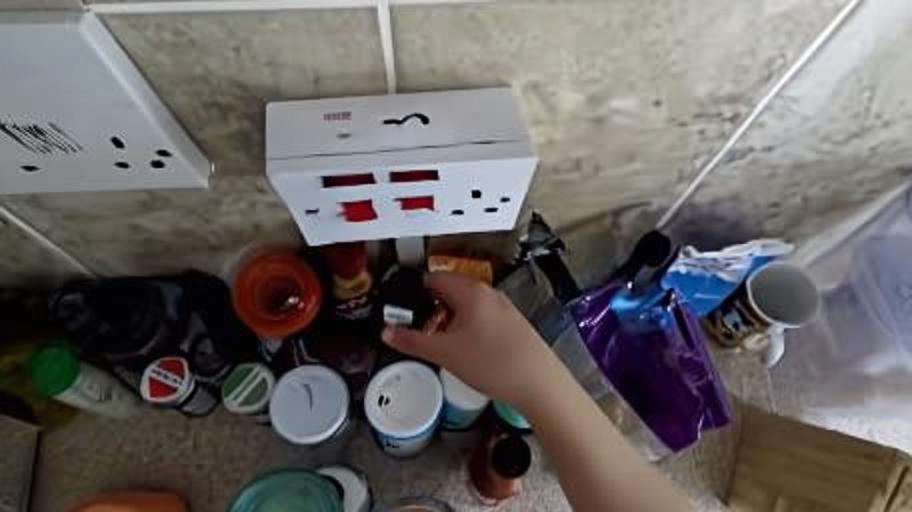}}
 &{\includegraphics[width=\linewidth]{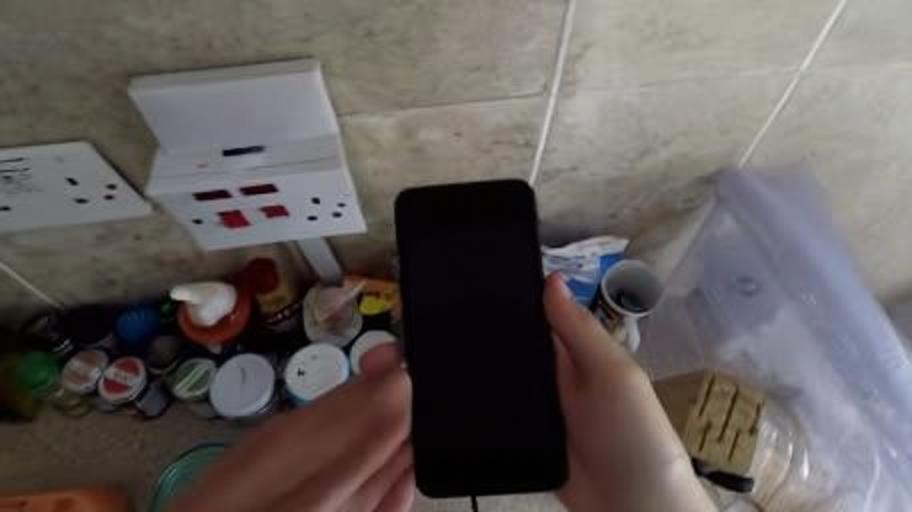}}
 & {\includegraphics[width=\linewidth]{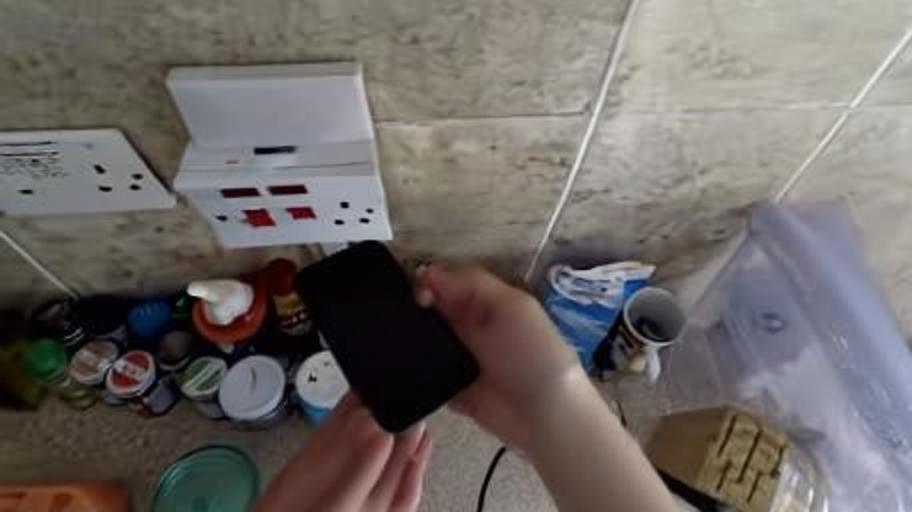}}
\\
\multicolumn{5}{p{\textwidth}}{\textbf{Forward prompt:} connect the black plug to the white socket. \newline\textbf{Reverse prompt:} Disconnect the black plug to the white socket.}\\
\midrule
{\includegraphics[width=\linewidth]{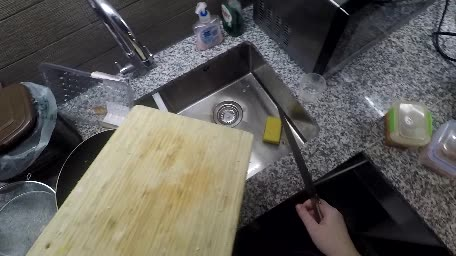}} & 
{\includegraphics[width=\linewidth]{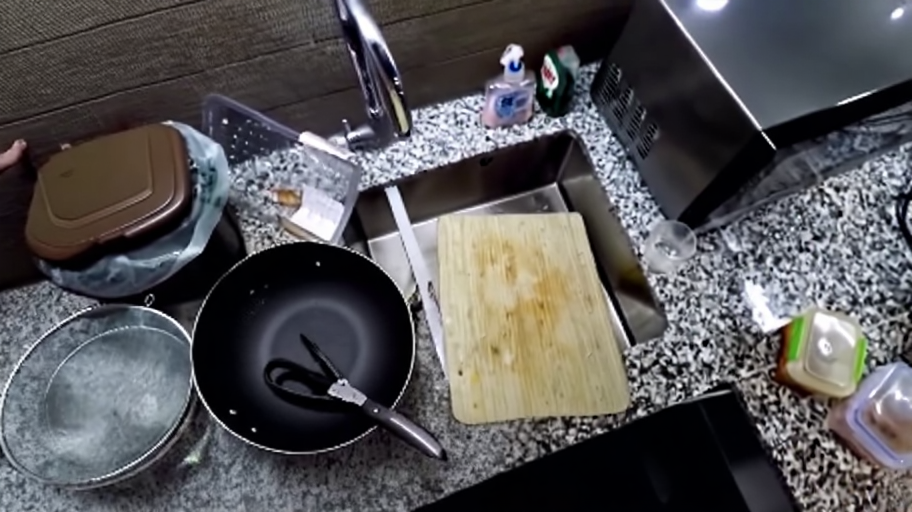}}
 &{\includegraphics[width=\linewidth]{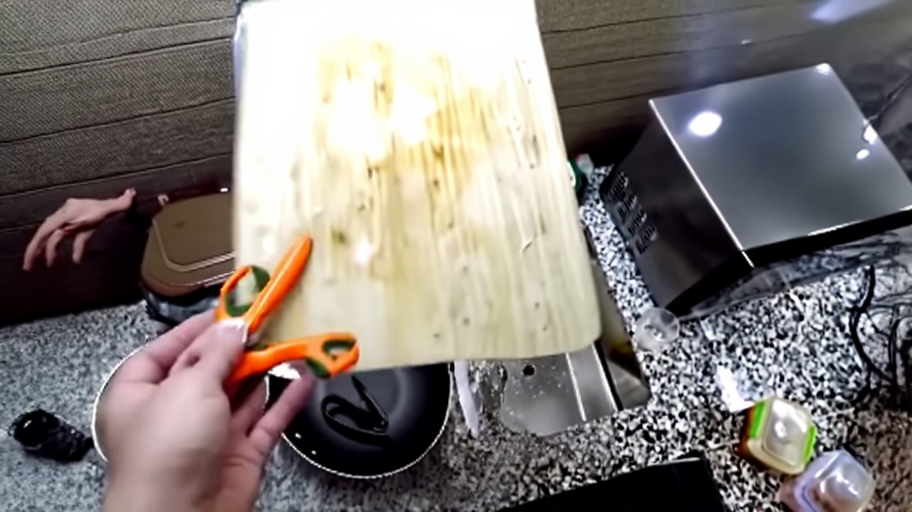}}
 &{\includegraphics[width=\linewidth]{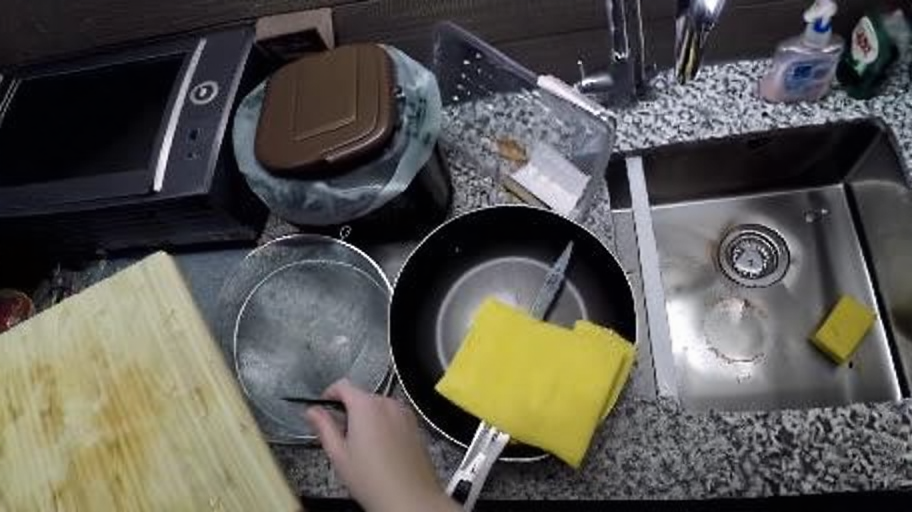}}
 & {\includegraphics[width=\linewidth]{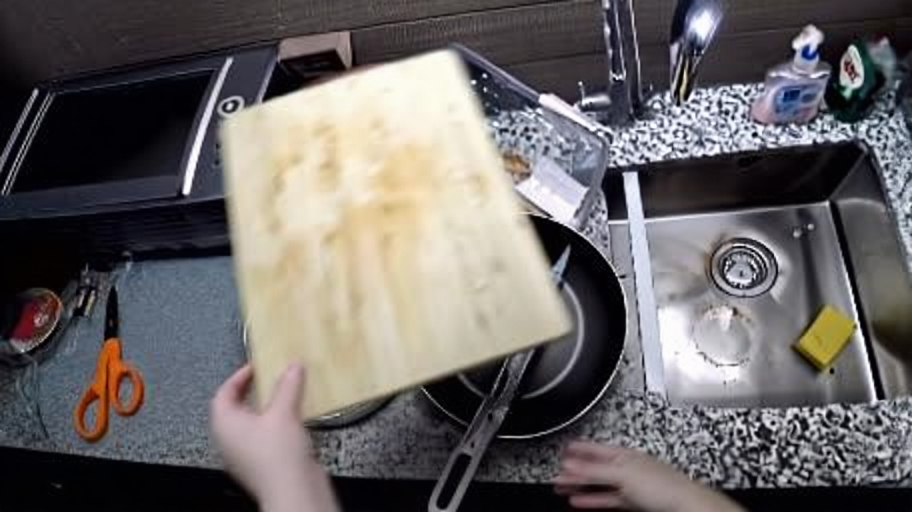}}
\\
\multicolumn{5}{p{\textwidth}}{\textbf{Forward prompt:} From an egocentric view,  a hand places a wooden chopping board onto a granite countertop beside a stainless steel sink. The board is positioned near a black frying pan with scissors, a yellow sponge, and a microwave. The scene is lit by overhead artificial light, reflecting off metallic surfaces. \newline
\textbf{Reverse prompt:} From an egocentric view, hands lift a worn wooden cutting board from a stainless steel sink, moving it away. The speckled granite countertop holds a black frying pan with orange-handled scissors, a soap dispenser, and plastic containers. Bright overhead lighting reflects off the metal surfaces as the board is removed.
}\\
\midrule
{\includegraphics[width=\linewidth]{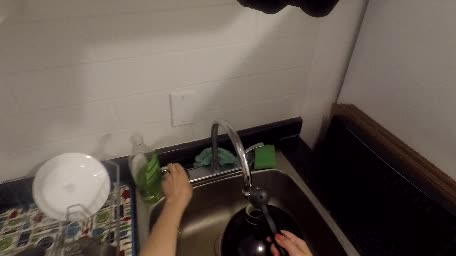}} & 
{\includegraphics[width=\linewidth]{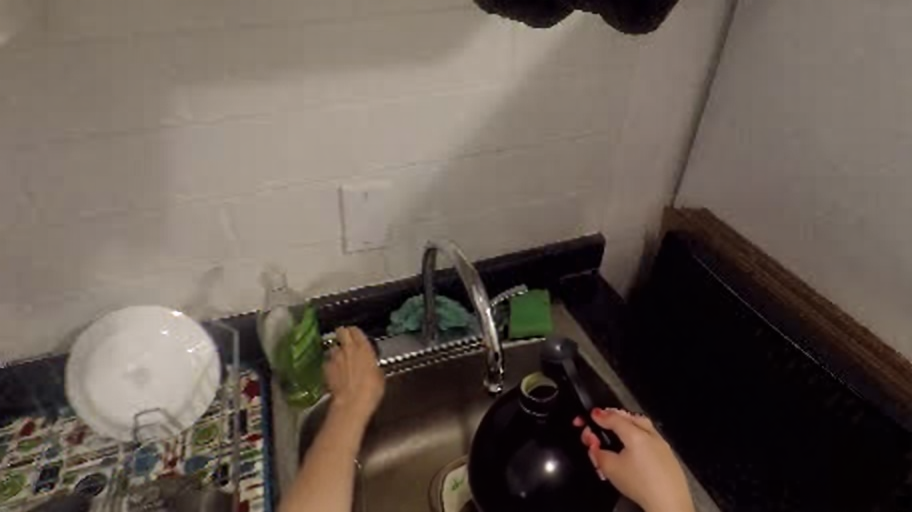}}
 &{\includegraphics[width=\linewidth]{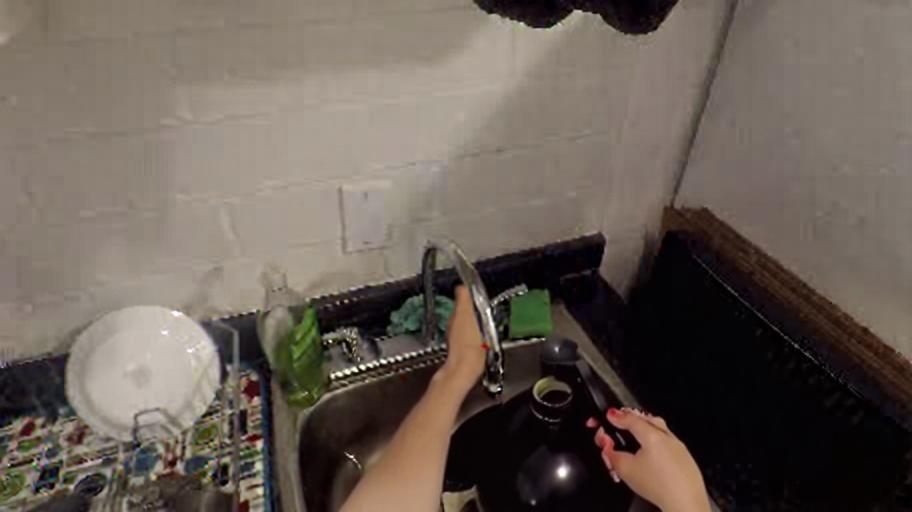}}
 &{\includegraphics[width=\linewidth]{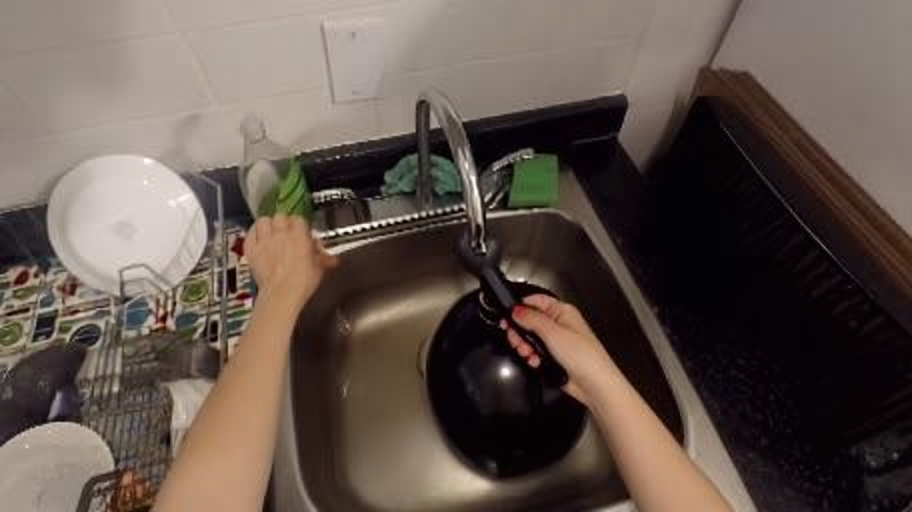}}
 & {\includegraphics[width=0.9\linewidth]{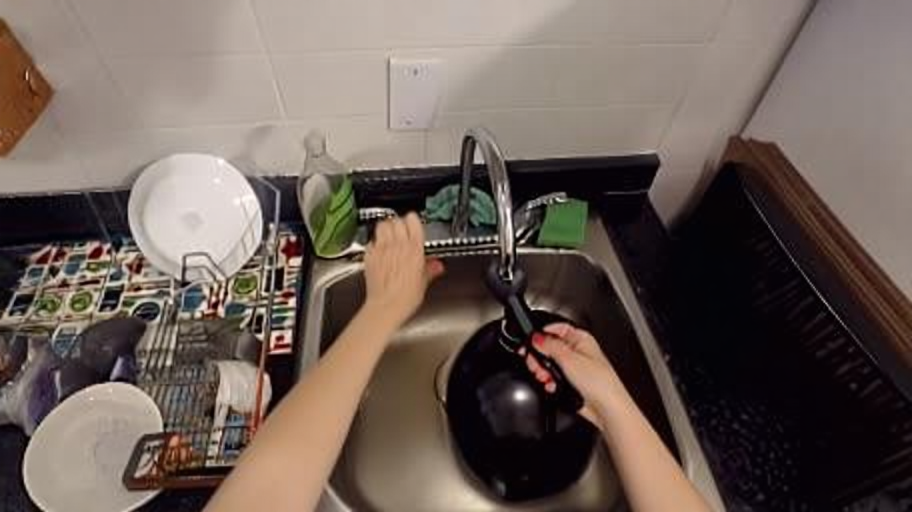}}
\\
\multicolumn{5}{p{\textwidth}}{\textbf{Forward prompt:} From a first-person view, a left hand turns the curved faucet handle to open it, while the right hand holds a dark pot under the stream. The scene is a dimly lit kitchen sink with a white tiled wall, a green sponge, dish soap, and a colorful dish rack with a white plate to the left.\newline
\textbf{Reverse prompt:} From an egocentric view, a left hand turns off a chrome faucet over a stainless steel sink, while a right hand holds a black pot. The scene includes a dish rack with a white plate, a green sponge, and a bottle on a black countertop, against white tiled walls under bright kitchen lighting.
}\\
\midrule
{\includegraphics[width=\linewidth]{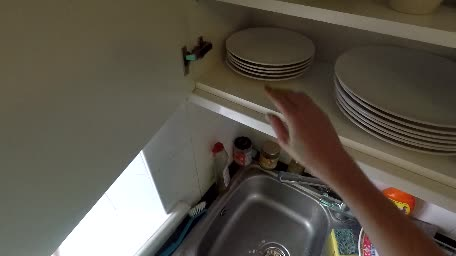}} & 
{\includegraphics[width=\linewidth]{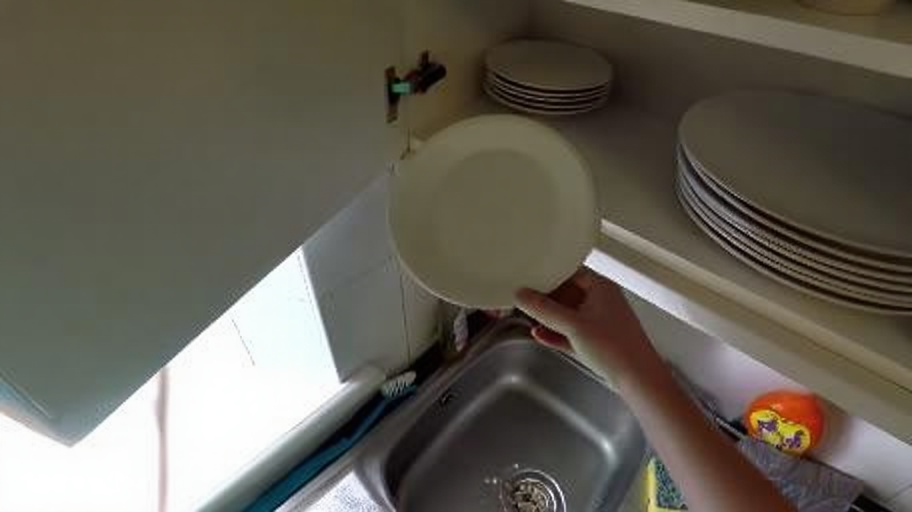}}
 &{\includegraphics[width=\linewidth]{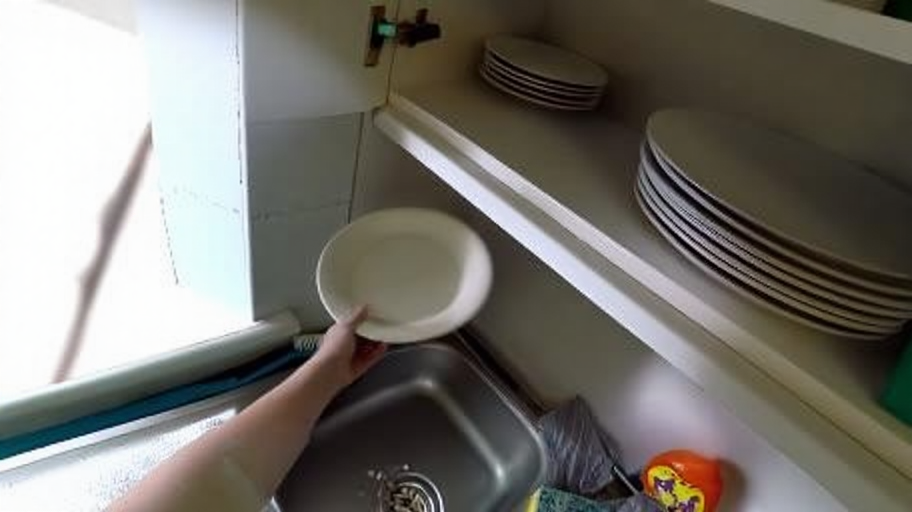}}
 &{\includegraphics[width=\linewidth]{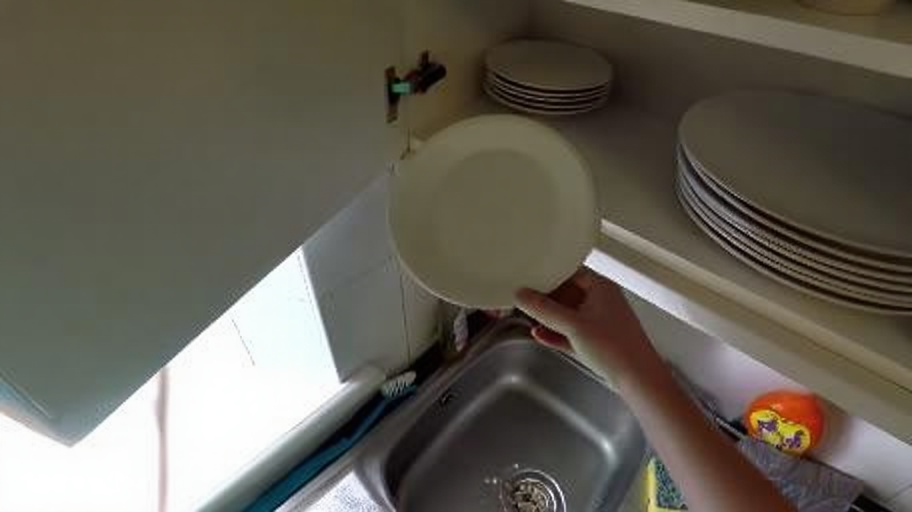}}
 & {\includegraphics[width=\linewidth]{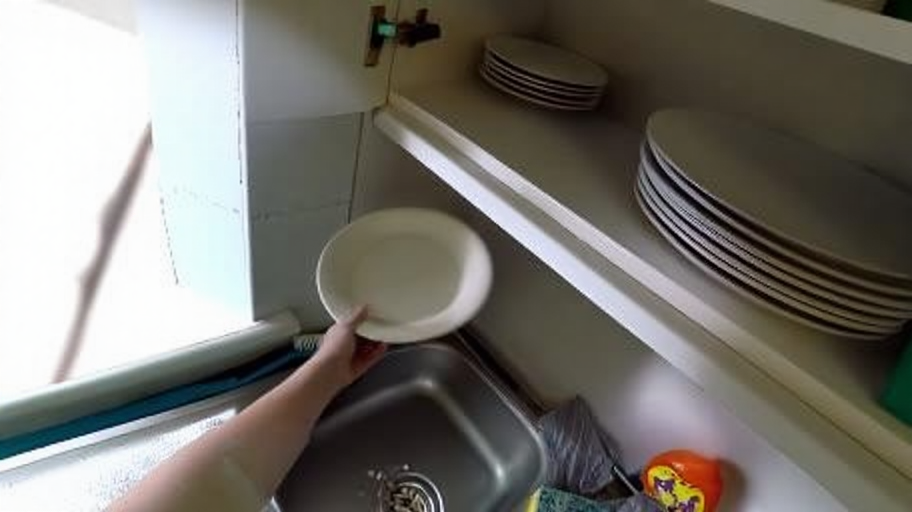}}
\\
\multicolumn{5}
{p{\textwidth}}{\textbf{Forward prompt:} From an egocentric view, a right hand reaches into an open white kitchen cabinet above a stainless steel sink, lifting a single white ceramic plate from a stack on the upper shelf, with natural light illuminating the scene from the left.\newline
\textbf{Reverse prompt:} From an egocentric view, a hand places a white ceramic plate onto a shelf inside an open white kitchen cabinet, next to two existing stacks of plates, above a stainless steel sink under bright overhead lighting.
}
\\ 
\bottomrule      
 \end{tabularx}
 \captionof{figure}{\textbf{Qualitative results.} Qualitative comparison BAGEL \cite{deng2025bagel} with BAGEL-DoUndo.} 
\label{fig:supp:qualitative}
\end{table*}



\end{document}